\documentclass[letterpaper]{article} 
\usepackage[]{aaai23}  
\usepackage{times}  
\usepackage{helvet}  
\usepackage{courier}  
\usepackage[hyphens]{url}  
\usepackage{graphicx} 
\usepackage{natbib}  
\usepackage{caption} 
\usepackage{subcaption}
\usepackage{booktabs}           
\usepackage{amsmath,amssymb,amsthm,textcomp}
\DeclareMathOperator*{\argmax}{arg\,max}

\newtheorem{definition}{Definition}
\usepackage{algorithm}
\usepackage{algorithmicx}
\usepackage{algpseudocode}

\usepackage{newfloat}
\usepackage{listings}
\DeclareCaptionStyle{ruled}{labelfont=normalfont,labelsep=colon,strut=off} 
\floatstyle{ruled}
\newfloat{listing}{tb}{lst}{}
\floatname{listing}{Listing}
\usepackage[disable]{todonotes}

\newcommand{\citeasnoun}[1]{\citet{#1}}

\title{Deceptive Reinforcement Learning in Model-Free Domains}
\author {
    Alan Lewis,\textsuperscript{\rm 1}
    Tim Miller, \textsuperscript{\rm 1}
}
\affiliations {
    \textsuperscript{\rm 1} The University of Melbourne\\
    lewisa1@student.unimelb.edu.au, tmiller@unimelb.edu.au
}

\usepackage{bibentry}

\begin{document}

\maketitle

\begin{abstract}
 This paper investigates deceptive reinforcement learning for privacy preservation in model-free and continuous action space domains. In reinforcement learning, the reward function defines the agent's objective. In adversarial scenarios, an agent may need to both maximise rewards and keep its reward function private from observers. Recent research presented the \emph{ambiguity model} (AM), which selects actions that are ambiguous over a set of possible reward functions, via pre-trained $Q$-functions. Despite promising results in model-based domains, our investigation shows that AM is ineffective in model-free domains due to misdirected state space exploration. It is also inefficient to train and inapplicable in continuous action space domains. We propose the \emph{deceptive exploration ambiguity model} (DEAM), which learns using the deceptive policy during training, leading to targeted exploration of the state space. DEAM is also applicable in continuous action spaces. We evaluate DEAM in discrete and continuous action space path planning environments. DEAM achieves similar performance to an optimal model-based version of AM and outperforms a model-free version of AM in terms of path cost, deceptiveness and training efficiency. These results extend to the continuous domain.
\end{abstract}

\section{Introduction} \label{sec:intro}

Model-free reinforcement learning (RL) has emerged as a powerful framework for solving a wide variety of tasks, from challenging games \citep{silver2017mastering} to the protein folding problem \citep{jumper2021alphafold}. RL agents learn by interacting with their environment, with the objective to maximise accumulated rewards over a trajectory. The \emph{reward function} is a succinct and robust definition of the task, from which one can deduce an agent's policy \citep{sutton2018reinforcement, ng2000algorithms}. There are cases where we want to preserve the privacy of a reward function \citep{ornik2018deception}. Consider a military commander moving troops. If they fail to keep the destination private, an adversary can ambush them. This creates a dual objective---to reach the destination efficiently while not revealing the destination for as long as possible. 
As goal-directed actions leak information, optimising for a reward-function while keeping it private is challenging.

Deception is a key mechanism for privacy preservation in artificial intelligence (AI) \citep{liu2021deceptive,savas2021deceptive,ornik2018deception,masters2017deceptive}. Deception is the distortion of a target's perception of reality by hiding truths and/or conveying falsities \citep{bell2003toward,whaley1982toward}. In the military example, the commander may hide the true destination by moving towards a fake destination. Although there are successful deceptive AI methods, most are model-based, requiring prior knowledge of the environment dynamics \citep{savas2021deceptive,kulkarni2018resource,masters2017deceptive}. 

Recently, \citeasnoun{liu2021deceptive}  introduced the \emph{ambiguity model} (AM) for deception in RL. AM selects actions that are ambiguous over a set of candidate reward functions---including the real reward function and at least one fake reward function. AM estimates the probability that a candidate is real from an observer's perspective using a pre-trained $Q$-function for each candidate. 
It then selects the action that maximises the entropy of this probability distribution. \citeasnoun{liu2021deceptive} learn these $Q$-functions using value iteration. 

\begin{figure}[htb]
\begin{subfigure}[t]{.5\linewidth}
    \centering
    \includegraphics[width=1.\linewidth]{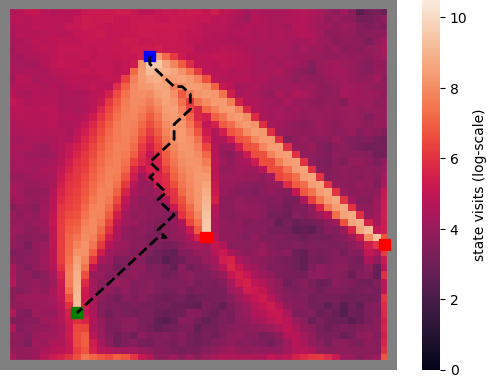}
    \subcaption[]{Model-free AM}
    \label{fig:AM state space exploration}
\end{subfigure}%
\begin{subfigure}[t]{.5\linewidth}
    \centering
    \includegraphics[width=1.\linewidth]{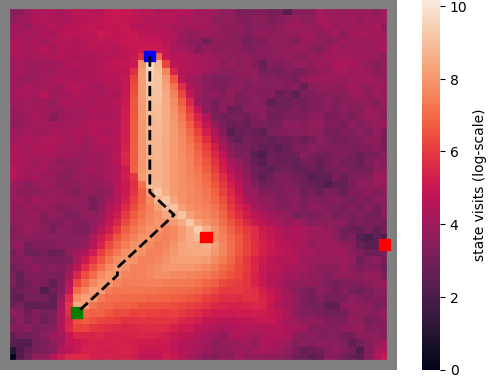}
    \subcaption[]{DEAM}
    \label{fig:DEAM state space exploration}
\end{subfigure}%

\begin{subfigure}[t]{.5\linewidth}
 \centering
 \includegraphics[width=1.\textwidth]{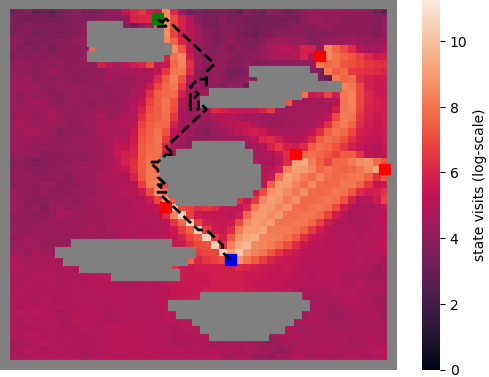}
 \subcaption[]{Model-free AM}
 \label{fig:AM inefficient action selection}
\end{subfigure}%
\begin{subfigure}[t]{.5\linewidth}
 \centering
 \includegraphics[width=1.\textwidth]{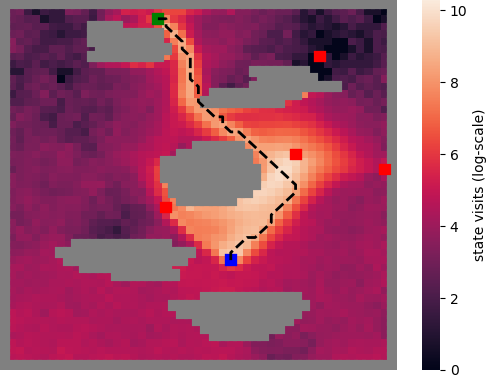}
 \subcaption[]{DEAM}
 \label{fig:DEAM inefficient action selection}
\end{subfigure}%
\caption[]{
State-visitation heatmaps vs deceptive path. \textit{Start:} blue. \textit{Real goal:} green. \textit{Fake goals:} red. \textit{Final path:} dashed line. Lighter parts denote more frequently visited states.
}
\label{fig:state space expoloration}
\end{figure}

We carefully study the AM, finding that it surprisingly fails in model-free domains---that is, using a model-free approach to learn $Q$-functions leads to poor performance. Our analysis shows why:  by separately pre-training honest subagents to learn $Q$-functions, AM explores the states along the honest trajectories `towards' each candidate reward function, instead of those visited by the deceptive policy. Figures~\ref{fig:AM state space exploration}~and~\ref{fig:AM inefficient action selection} show AM's state space exploration in a path planning domain (lighter parts denote more frequently visited states). There is little overlap between the heavily explored states and the final deceptive path; showing that AM over-explores states that have little influence on its final policy, and under-explores states along good deceptive trajectories. This leads to inaccurate $Q$-values for these states and a poor policy. AM's ineffectiveness with model-free methods prevents its application to many problems, including model-free and simulation-based problems.

We propose the \emph{deceptive exploration ambiguity model} (DEAM), an ambiguity-based model that collectively learns candidate $Q$-functions using the entropy-based policy to select actions during training. DEAM has three key improvements compared to AM. 
First, by training with a deceptive policy, DEAM explores states along deceptive trajectories, shown by the overlap between the frequently visited states and the final path in Figures~\ref{fig:DEAM state space exploration}~and~\ref{fig:DEAM inefficient action selection}. This leads to effective deceptive behaviour in model-free domains.
Second, DEAM extends to continuous action space domains by using actor-critic (AC) subagents to learn $Q$-functions and changes to  action selection. Third, DEAM shares experiences between subagents, improving training efficiency as all subagents learn from every environment interaction.

We evaluate DEAM in discrete and continuous path planning domains, assessing deceptiveness, path costs and training efficiency. DEAM is more deceptive than an honest agent, with similar performance to an optimal, model-based (value iteration) version of AM. DEAM resolves the misdirected state space exploration problem, leading to a more cost-efficient policy, reaching the real goal in $1.1\times$ the optimal path cost compared to $1.6\times$ for the model-free version of AM. DEAM converges to a deceptive policy in approximately $70\%$ of the allotted training steps, while the model-free AM does not converge in the allotted interactions.

Overall, we make three key contributions to deceptive RL:
\begin{enumerate}
    \item Via a thorough analysis, we identify important weaknesses in AM that are not apparent in \citeasnoun{liu2021deceptive}'s paper. These weaknesses make AM ineffective with model-free methods, limiting its application.
    \item We resolve these weaknesses by introducing DEAM, the only model-free deceptive RL framework. This enables
    a significant expansion in the scope of deceptive RL, including environments with unknown dynamics.
    \item By evaluating the agents against passive and active adversaries, we show scenarios where DEAM and AM's entropy-based model leads to counter-productive behaviour, pointing to important areas for future research.
\end{enumerate}

\section{Related Work}
\label{sec:lit}

\citeasnoun{masters2017deceptive} define deception through \emph{simulation} and \emph{dissimulation} \cite{bell2003toward, whaley1982toward}. We focus solely on dissimulation. Dissimulation measures the entropy \citep{shannon1948mathematical} over the candidate goal probabilities: 
$
  \textrm{dissimulation} = -\sum_{g_i \in G} P(g_i \mid \vec{o}) \cdot \log_2{P(g_i \mid \vec{o})}
$,
where $G$ is a set of candidate goals and $P(g_i\mid\vec{o})$ is the probability that $g_i$ is real from an observer's perspective, given the observation sequence $\vec{o}$. This definition is measurable, enabling agents to manage the trade-off between deception and rewards. However, their deceptive agent requires deterministic environments with known dynamics. DEAM is applicable in model-free, stochastic and continuous environments.

\citeasnoun{ornik2018deception} extend deception to stochastic environments with a deceptive Markov Decision Process (MDP), which incorporates the observer's beliefs into a regular MDP. Section~\ref{sec:prelim} has a precise definition. They highlight several challenges to optimisation in a deceptive MDP, such as a lack of knowledge of the observer's beliefs and unifying 
beliefs with environment rewards. \citeasnoun{savas2021deceptive} use an intention recognition system to model the observer's beliefs. Despite effectively 
balancing deception and rewards, their approach is model-based.

\citeasnoun{liu2021deceptive}'s AM for deceptive RL is closest to ours. Although it only requires $Q$-functions, and therefore theoretically extends to model-free domains, they use value iteration to learn these $Q$-functions in their evaluation. We show that AM is ineffective when using model-free RL, due to misdirected state space exploration. Our model resolves this issue by following the deceptive policy during training. 
We also enable application in continuous action spaces.


Finally, intention and plan recognition research is relevant since it is an inverse problem to deception. Plan recognition is the task of inferring an agent's plan by observing its behaviour \citep{ramirez2009plan,masters2019goal}. A common approach is cost-based plan recognition, which measures the divergence of behaviour from optimal. Intuitively, a rational agent will take a cost-minimising path. Therefore, the goal that best explains the observed path is more likely to be the real goal. This can be used to generate a probability distribution over the candidate goal-set \citep{masters2017cost}. \emph{Inverse reinforcement learning} (IRL) is the task of deducing a reward function given traces of the agent's behaviour \citep{ng2000algorithms}. However, IRL requires many behaviour traces whereas deceptive RL aims to deceive over a single trace of behaviour.

\section{Preliminaries} \label{sec:prelim}
\begin{definition}[Markov decision process (MDP) \citep{puterman2014markov}] \label{Markov Decision Process}
An MDP is defined by a tuple $(\mathcal{S},\mathcal{A},\mathcal{T},r,\gamma)$, where $\mathcal{S}$ is the state space, $\mathcal{A}$ is the action space, $\mathcal{T}(s_t, a_t, s_{t+1})$ is a transition function defining the probability of moving from state $s_t$ to state $s_{t+1}$ given action $a_t$, $r(s_t, a_t, s_{t+1})$ is the reward received for executing action $a_t$ in state $s_t$ and transitioning to state $s_{t+1}$, and $\gamma \in (0, 1)$ is a discount factor. 
\end{definition}

An MDP is solved by a \emph{decision policy} $\pi: \mathcal{S} \rightarrow \mathcal{A}$, which selects actions given a state. The objective is to maximise the value function: 
$
    V_\pi(s) = \mathbb{E}[\sum_{t\in T} \gamma^t r(s_t, a_t, s_{t+1})]\,.
$

\textbf{RL approaches.}
We separate RL approaches into three groups of methods: (1) \emph{Value-based}, such as \emph{$Q$-learning}; (2)~\emph{Policy gradient (PG)}; and (3) \emph{AC}. Value-based methods learn a $Q$-function $Q: \mathcal{S} \times \mathcal{A} \rightarrow \mathbb{R}$, which maps all state-action pairs onto an estimate for the true value function \citep{sutton2018reinforcement}; $Q(s_t,a_t)$ represents the future expected reward from selecting action $a_t$ in state $s_t$ and following $\pi$ from there. $Q$-learning agents iterate over the action space to maximise $Q$-values, which is infeasible in continuous action spaces \citep{sutton2018reinforcement}. PG methods learn a policy $\pi : \mathcal{S} \rightarrow \mathcal{A}$ using gradient descent \citep{sutton2018reinforcement}, so are applicable in continuous action spaces. However, they suffer from high variance and instability \citep{sutton1999policy}. AC methods combine PG methods with value-based  methods \citep{konda2000actor}. The actor is policy-based, enabling application in continuous action spaces. The critic learns the value function for stability. 

\textbf{Deceptive RL.}
\citeasnoun{ornik2018deception} define a \emph{deceptive MDP}. As agents need to maximise rewards and deceive observers, they include the observer's beliefs in the task:

\begin{definition}[Deceptive MDP] \label{def:Deceptive MDP}
A deceptive MDP is defined by a tuple $(\mathcal{S}, \mathcal{A}, \mathcal{T}, \mathcal{R}, r, \mathcal{B},\mathcal{L},\gamma)$, where $\mathcal{S}$, $\mathcal{A}$, $\mathcal{T}$, $r$, and $\gamma$ are the same as a regular MDP, $\mathcal{R}$ is a set of candidate reward functions, including the real reward function $r$ and at least one fake reward function, $\mathcal{B}$ is the observer's belief-set and $\mathcal{L}(s_t, a_t, s_{t+1}, b_t)$ is a belief-induced reward function, defining the reward for executing action $a_t$ in state $s_t$ and transitioning to state $s_{t+1}$ incorporating the impact of the observer's belief $b_t$ about the agent's real reward function.
\end{definition}

\textbf{Deceptive RL challenges.} 
The deceptive RL objective is to maximise the belief-induced value function: 
$
V_\pi(s) = \mathbb{E}[\sum_{t\in T} \gamma^t \mathcal{L}(s_t, a_t, s_{t+1}, b_t)]\,.
$
Compared to standard RL, there are three challenges. First, as $\mathcal{L}$ is abstract in Definition~\ref{def:Deceptive MDP}, we need to instantiate $\mathcal{L}$ such that it balances the trade-off between deception and reward-maximisation. Since there are two competing objectives, we can frame this as a \emph{multi-objective RL} (MORL) problem; given an observer model and a mapping from observer beliefs to a deceptiveness score $\mathcal{D}(b_t):\mathcal{B} \rightarrow \mathbb{R}$, we can define $\mathcal{L}$ as a vector of the two objectives: $\mathcal{L}(s_t, a_t, s_{t+1}, b_t) = \langle r(s_t, a_t, s_{t+1}),  \mathcal{D}(b_t) \rangle$. This is useful as MORL offers established optimisation methods. Second, we need to model observer beliefs $b_t$. We can conceptualise $b_t$ as a probability distribution where probabilities are the likelihood that each candidate reward function is real \citep{liu2021deceptive,ornik2018deception,savas2021deceptive}. This transforms the challenge to an intention recognition problem. Third, we need to define a deceptiveness mapping $\mathcal{D}$. Recent approaches use simulation and dissimulation \cite{liu2021deceptive, savas2021deceptive, nichols2022adversarial}.  

\section{Ambiguity Model} \label{sec:AM}
\citeasnoun{liu2021deceptive} introduce AM for dissimulative deception in RL settings. 
Below we outline AM in the context of how it resolves the three deceptive RL challenges.  

\textbf{Modelling observer beliefs.} AM models observer beliefs via cost-based plan recognition. Rather than using cost-differences to measure the divergence of behaviour from optimal, it uses \emph{$Q$-differences}:
\begin{equation} \label{Q difference}
    \Delta_{r_i}(\vec{o}) = \sum_{(s_t,a_t) \in \vec{o}}(Q_{r_i}(s_t,a_t) - \max_{a_t' \in A} Q_{r_i}(s_t,a_t'))\,.
\end{equation}
If behaviour is optimal for a reward function $r_i$, then $\Delta_{r_i}(\vec{o})=0$. Otherwise, $\Delta_{r_i}(\vec{o})<0$ and decreases with  more sub-optimality. AM estimates probabilities for the candidate reward functions via a Boltzmann distribution:
\[
    P(r_i\mid\vec{o})= \frac{\exp(\Delta_{r_i}(\vec{o}))}{\sum_{r_j \in \mathcal{R}}\exp(\Delta_{r_j}(\vec{o}))} \cdot P(r_i)\,,
\]
where $P(r_i)$ is an estimate of the prior probability that $r_i$ is the real reward function. 
A less optimal sequence leads to a lower probability. This is a proxy for the observer beliefs.

\textbf{Deceptiveness mapping.} AM uses dissimulation (entropy) to map observer beliefs to deceptiveness. Dissimulation makes actions ambiguous and therefore more resilient against any observer; actions that move equally towards two reward functions makes those reward functions indistinguishable, regardless of one's recognition system. This leads to deceptive behaviour against both na\"{i}ve plan recognition models and non-na\"{i}ve human observers \cite{liu2021deceptive}.

\textbf{Defining $\mathcal{L}$.} 
Instead of explicitly defining $\mathcal{L}$, AM satisfies the dual objective by maximising deceptiveness from a set of pruned actions that sufficiently satisfy the reward objective:
\begin{equation*}\begin{split}
    \pi(\vec{o}, s_t) = \argmax_{a \in A^+} -\sum_{r_i \in \mathcal{R}} &P(r_i\mid \vec{o} \cup (s_t, a)) \\
    &\cdot \log_2 P(r_i\mid\vec{o} \cup (s_t, a))\,,
\end{split}\end{equation*}
where $\vec{o} \cup (s_t, a)$ is the current observation sequence including $(s_t, a)$ and $A^+$ is the set of actions remaining after pruning. AM prunes actions according to:
\begin{equation} \label{equation:pruning}
\textrm{prune}(\delta, s_t) = \{a \in \mathcal{A}(s_t) \mid Q_{\textrm{real}}(s_t, a) - R > \delta\}\,,
\end{equation}
where $R=Q_{\textrm{real}}(s', a') - Q_{\textrm{real}}(s_0, a_0)$, $Q_{\textrm{real}}$ is the $Q$-function for the real reward function, and $(s', a')$ and $(s_0, a_0)$ are the previous and initial observations respectively. Thus, $R$ represents the \emph{residual} expected reward received so far. As such, actions that do not increase expected future rewards beyond the threshold $\delta$ are pruned from consideration. For path planning, this ensures that AM reaches the real goal. 
From the remaining actions, AM minimises the information gain for the observing agent. 
This mirrors \emph{multi-criteria RL}, an instance of MORL that maximises a single objective subject to the constraint that the other objectives meet a given threshold \cite{gabor1998multi}. Here, deceptiveness is the unconstrained objective and environment rewards is the constrained objective.

\textbf{Limitations.} First, AM learns $Q$-functions by separately pre-training a subagent for each candidate reward function. As discussed in Section~\ref{sec:intro}, this is ineffective in a model-free setting due to misdirected exploration of the state space. That is, AM explores the states along the honest trajectories to each candidate reward function, rather than those visited by the final policy, leading to poor deceptive behaviour. This restricts the application of AM to domains where state space exploration is not important, such as those with known environment models. Second, AM is inapplicable in continuous action spaces. In particular, the $Q$-difference formulation requires a maximisation over the action space and the pruning procedure considers all actions in the action space.

\section{Deceptive Exploration Ambiguity Model} \label{sec:DEAM}
\begin{figure*}[t]
    \centering
    \includegraphics[{width=0.9\linewidth}]{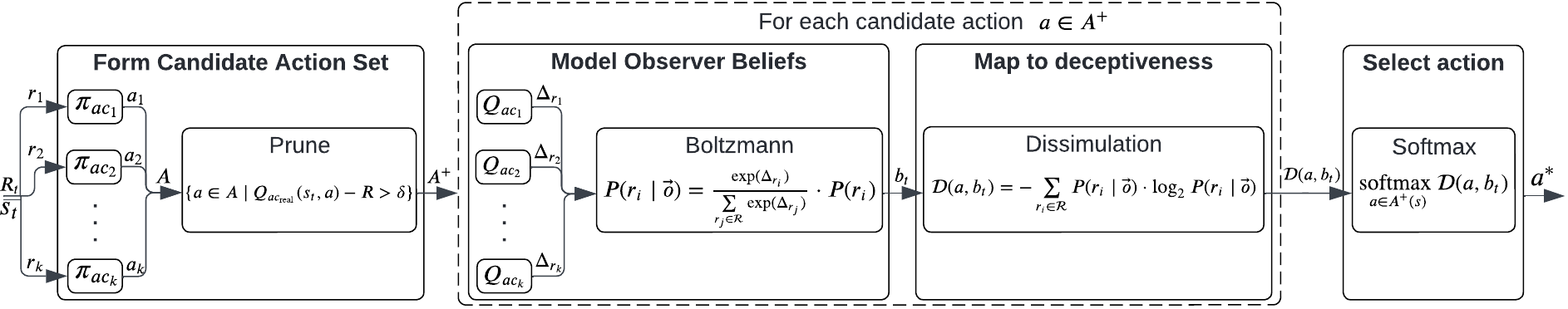}
    \caption{DEAM: DEAM forms a candidate action set using AC subagents, pruning those that insufficiently satisfy rewards. It selects the action that soft-maximises the entropy of the candidate reward probabilities from the modelled observer perspective.}
    \label{fig:DEAM}
\end{figure*}
We introduce DEAM to address these limitations with three improvements: (1) It explores deceptive trajectories during training, improving performance in model-free settings; (2) It shares experiences between subagents, improving training efficiency; and (3) It extends to continuous action spaces.

Figure \ref{fig:DEAM} shows DEAM's architecture and Algorithm~\ref{algorithm:DEAM} has the complete details. DEAM collectively trains AC subagents for each candidate reward function, using the entropy-based policy to select actions. Given a state $s_t$, each subagent uses its policy $\pi_{ac_i}$ to submit a candidate action $a_i$, forming a set of candidate actions $A$ (line~\ref{alg:line 5}). From these actions, DEAM prunes those that lead to insufficient rewards with the same pruning procedure as AM (Equation~\ref{equation:pruning}), leading to a pruned action set $A^+$ (line~\ref{alg:line 6}). For each of these actions, DEAM calculates the entropy of the reward function probabilities using an observer model (lines~\ref{alg:line 9} and \ref{alg:line 10}). It then selects the action $a^*$ that soft-maximises entropy (line~\ref{alg:line 12}), using a temperature parameter $\tau$ which decays during training (line~\ref{alg:line 16}). This leads to an environment update and a reward for each candidate reward function. The experience $(s_t, a^*, s_{t+1}, r_i(s_t, a^*, s_{t+1}))$ is added to the replay buffer of each subagent (line~\ref{alg:line 13}) which improves policies and $Q$-functions at the end of each epoch (line~\ref{alg:line 17}). Below we outline DEAM's three improvements.

\begin{algorithm}[t]
\begin{algorithmic}[1]
\Require{$k$ candidate reward functions}
\Require{prior probability $p_i$ for each reward function}
\Require{soft-max temperature ($\tau$), $\tau$ decay rate ($\lambda_{\tau}$), and pruning threshold ($\delta$)}
\State{$\mathcal{AC} \leftarrow \{ac_1,...,ac_k\}$ \Comment{Instantiate AC subagents and}} \State{$\mathcal{D} \leftarrow \{D_1,...,D_k\}$ \Comment{Instantiate replay buffers}}
\For{each training episode}
    \State{$\vec{o} \leftarrow$  \{\}} 
    \Comment{Initialise observation sequence}
    \For{each environment step ($s_t$)}
        \State{$A \leftarrow \{a_i \mid a_i \sim \pi_{ac_i}(s_t) \textrm{ for } ac_i \textrm{ in } \mathcal{AC}\}$}
        \label{alg:line 5}
        \State{$A^+ \leftarrow$ prune($A$, $\delta$)} \label{alg:line 6}
        \State{$E \leftarrow \{\}$}
        \For{$a \in A^+$}
            \State{$\vec{o_a} \leftarrow \vec{o} \cup (s_t, a)$}
            \State{$P \leftarrow \{p'_i | p'_i = \frac{\exp(\Delta_{r_i} (\vec{o_a} )) \cdot p_i}{\underset{r_j \in \mathcal{R}}{\sum} \exp\Delta_{r_j}(\vec{o_a})} \textrm{ for } r_i \in \mathcal{R} \}$} \label{alg:line 9}
            \State{$E \leftarrow E \cup \{(a, \textrm{entropy}(P))\}$} \label{alg:line 10}
        \EndFor{}
        \State{$a^* \leftarrow \textrm{softmax}(E, \tau)$} \label{alg:line 12}
        \State{$D_i \leftarrow D_i \cup \{(s_t, a^*, s_{t+1}, r_i\} \textrm{ for } D_i \in \mathcal{D}$} \label{alg:line 13}
        \State{$\vec{o} \leftarrow \vec{o} \cup (s_t, a^*)$}
    \EndFor{}
    \State{$\tau \leftarrow \tau \cdot \lambda_{\tau}$} \Comment{Decay $\tau$} \label{alg:line 16}
    \State{Update subagents using replay-buffers in $\mathcal{D}$} \label{alg:line 17}
\EndFor{}
\caption{Deceptive Exploration Ambiguity Model}
\label{algorithm:DEAM}
\end{algorithmic}
\end{algorithm}

\textbf{Deceptive exploration during training.}
Rather than separately pre-training an honest subagent for each candidate reward function, DEAM trains the subagents together, using the deceptive policy to select actions. As such, DEAM explores the deceptive part of the state space, shown by the overlap in the heavily visited states and the deceptive path in Figures~\ref{fig:DEAM state space exploration} and \ref{fig:DEAM inefficient action selection}. Thus, it uses training time to learn a policy and $Q$-functions for the states visited by the final policy.

To accommodate collectively training subagents using the deceptive policy, DEAM uses a soft-maximum policy that converges to a hard-maximum policy as training progresses. During early training, $Q$-functions are inaccurate. When using a hard-maximum policy, this inaccuracy favours actions submitted by subagents with high magnitude $Q$-functions. This occurs because high magnitude $Q$-functions lead to relatively high $Q$-differences for sub-optimal actions (i.e. those submitted by the other subagents). This leads to a skewed probability distribution for these actions. Actions submitted by subagents with high $Q$-functions are approximately optimal for themselves, avoiding the disproportionately high $Q$-difference and skewed distribution. As such, these actions tend to have higher entropy values and are selected disproportionately often. This leads to positive rewards for these subagents, compounding the problem. The soft-maximum policy selects actions with probability:
$\label{soft-max policy}
    P(a_i) = \exp(e_i/\tau)/\sum_{e_j \in E}\exp(e_j/\tau)$,
where $e_i$ is the entropy value of action $a_i$. The temperature parameter $\tau$ starts high, leading to more randomness and therefore exploration. This allows all subagents to learn more accurate $Q$-values before exploiting the deceptive policy as $\tau$ decays.


\textbf{Sharing experiences between subagents.}
As AM separately pre-trains subagents, they do not learn from the experiences sampled by the other subagents. DEAM's subagents learn from all environment interactions, even if the selected action is submitted by another subagent. That is, DEAM adds $(s_t, a^*, s_{t+1}, r_i(s_t, a^*, s_{t+1}))$ to the replay buffer of every subagent (line \ref{alg:line 13}). This improves training efficiency as subagents learn from many experiences without the cost of sampling them. For subagents to learn from experiences sampled by other subagents, they must use \emph{off-policy} RL 
\citep{sutton2018reinforcement}. 
AM can also benefit from shared experiences. But, it does not solve the misdirected exploration that leads to poor performance in model-free domains.

\textbf{Extension to continuous action spaces.}
We make three changes to enable application in continuous action space domains. First, we use AC subagents to learn $Q$-functions. The policy-based actor submits actions while the $Q$-learning critic calculates $Q$-differences, leading to probabilities for the candidate reward functions. Second, instead of using the $Q$-difference calculation, as per Equation~\ref{Q difference}, DEAM uses the action selected by the actor as the optimal action and the $Q$-function of the critic to evaluate that state-action pair. This leads to a modified $Q$-difference formula:
\[
    \Delta_{r_i}(\vec{o}) = \sum_{(s,a) \in \vec{o}}(Q_{r_i}(s,a) - Q_{r_i}(s,a^*)) \mbox{ where } a^* \sim \pi_i(s).
\]
As the policy is stochastic, $\max_{a \in A} Q_{r_i}(s,a) \neq Q_{r_i}(s,a^*)$, even for a converged subagent. However, as $\max_{a \in A} Q_{r_i}(s,a)$ is intractable, $Q_{r_i}(s,a^*)$ is an appropriate approximation. Third, instead of selecting the optimal action from all actions in the action space, DEAM selects from a candidate action set formed using subagent policies. This supports continuous action spaces as the size of the candidate action set is equal to the number of candidate reward functions rather than the size of the action space. However, it can lead to suboptimal action selection. The subagents may not submit the optimal action to the candidate action set and therefore DEAM would not select this action. Despite the theoretical suboptimality, we show that DEAM performs as well as an optimal value iteration AM, that does not have this problem, in the discrete domain. Also, restricting the size of the candidate action set is required to extend the model to continuous action spaces.

\section{Evaluation} \label{sec:evaluation}
We conduct two experiments. The first uses a passive observer to estimate a probability distribution over the candidate goals without participation in the environment. This follows \citeasnoun{liu2021deceptive}'s experimental set-up. The second evaluates the agents using an active adversary that needs to beat the agent to the real goal. This follows \citeasnoun{nichols2022adversarial}'s \emph{pirate deception scenario}. Our experiments aim to assess whether DEAM addresses the limitations of AM. 

\textbf{Independent variables.} We compare DEAM with: (1) an honest agent (moves directly to the real goal using $Q_{\textrm{real}}$); (2) an optimal, model-based AM (denoted: value iteration AM); and (3) a model-free, AC version of AM (denoted: AM).
We do not know of other model-free RL agents to use as baselines.
Value iteration AM is the upper bound possible for DEAM as it calculates dissimulation optimally. For the model-free AM, we add the changes that enable extension to continuous action spaces to allow evaluation in the continuous domain.

\subsection{Passive Observer Experiment Design} \label{sec:experiment design}

\textbf{Environments.} 
We assess the agents in two path planning environments: one discrete and one continuous. The discrete environment supports lateral and diagonal movements. Agents receive a reward of $100$ for reaching the goal, $-1$ for lateral actions and $-\sqrt{2}$ for diagonal actions. We use the same 40 environment settings (5 maps $\times$ 8 goal placements) as \citeasnoun{liu2021deceptive}. The maps have varied sizes and obstacles as follows: (1) $49\times49$ without obstacles, (2) $49\times49$ with few large obstacles, (3) $49\times49$ with many small obstacles, (4) $100\times100$ with large `island' obstacles, and (5) $100\times100$ maze. In the continuous environment \citep{henderson2017multitask}, agents select a velocity between $[-1, 1]$ and direction between $[0, 2\pi]$. Rewards are based on the distance to the goal. We use the first 4 maps from the discrete environment, giving a total of 32 settings. We omitted the maze map since it was too time consuming to train even an honest agent. Value iteration AM is inapplicable in continuous domains, so it is omitted from the second environment.

\textbf{Agent implementation.} We use \citeasnoun{haarnoja2018soft}'s soft actor-critic (SAC) as the subagent model. Since SAC is an \emph{off-policy} algorithm, it enables shared experiences between subagents. Hyperparameter choices are taken from the original SAC paper. We found that higher stochasticity was beneficial during the start of training and less so as training progressed. Therefore we chose $\tau=1$ and $\lambda_{\tau}=0.9$. We chose $\delta=0$ since we found that aggressive pruning is required to reliably reach the real goal (see Section~\ref{sec:hyperparameter studies}).

\textbf{Measures.} For deceptiveness, we measure the real goal probability at each time-step, using \citeasnoun{masters2017cost} intention recognition model. We also measure the number of steps taken after the last deceptive point (LDP). 
The LDP is the point after which the real goal is most likely from the observer's perspective \citep{masters2017deceptive}. After the LDP, a rational observer is not deceived. Thus, fewer steps after the LDP indicates better deception. For reward, we use path cost, proportional to the optimal, honest path cost. Finally, we use the number of environment steps to convergence for training efficiency.

\subsection{Passive Observer Experiment Results} \label{sec:results}

\begin{figure*}[t]
\centering
\begin{subfigure}[t]{.275\textwidth}
    \centering
    \includegraphics[width=\textwidth]{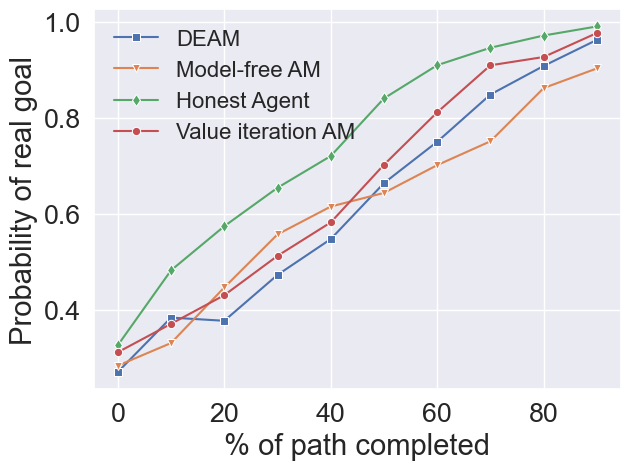}
    \subcaption[]{Deceptiveness (discrete)}
    \label{fig:discrete rg_probs}
\end{subfigure}%
\begin{subfigure}[t]{.275\textwidth}
    \centering
    \includegraphics[width=\textwidth]{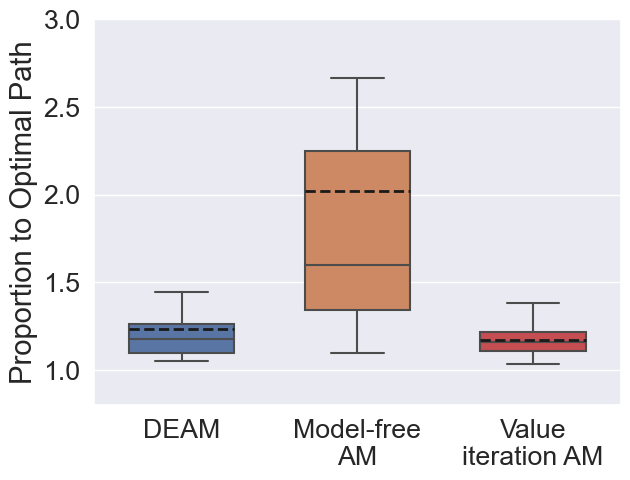}
    \subcaption[]{Path costs (discrete)}
    \label{fig:discrete path_cost}
\end{subfigure}%
\begin{subfigure}[t]{.275\textwidth}
    \centering
    \includegraphics[width=\textwidth]{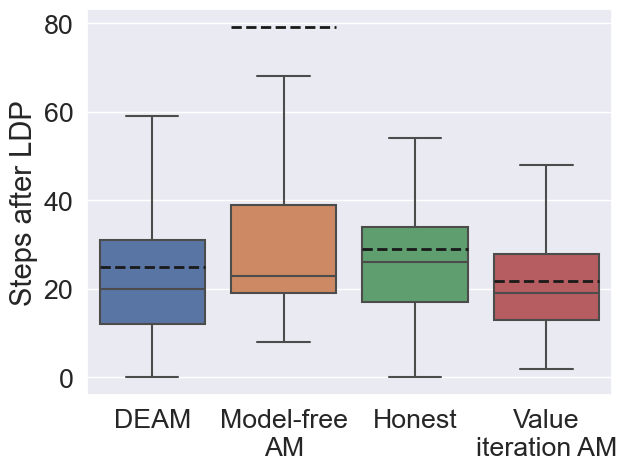}
    \subcaption[]{Steps after LDP (discrete)}
    \label{ldp}
\end{subfigure}%

\begin{subfigure}[t]{.275\textwidth}
    \centering
    \includegraphics[width=\textwidth]{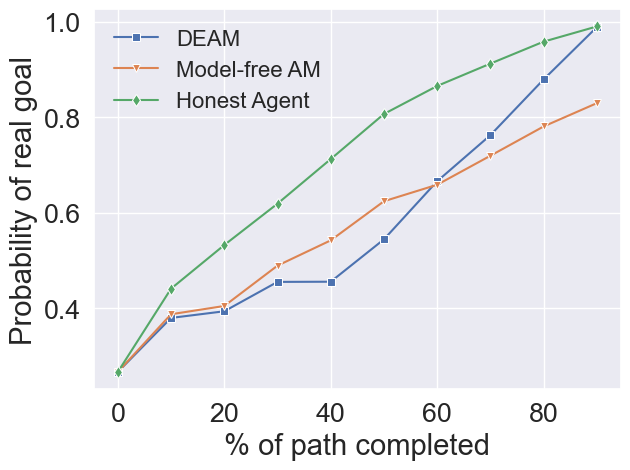}
    \subcaption[]{Deceptiveness (continuous)}
    \label{fig:continuous rg_probs}
\end{subfigure}%
\begin{subfigure}[t]{.275\textwidth}
    \centering
    \includegraphics[width=\textwidth]{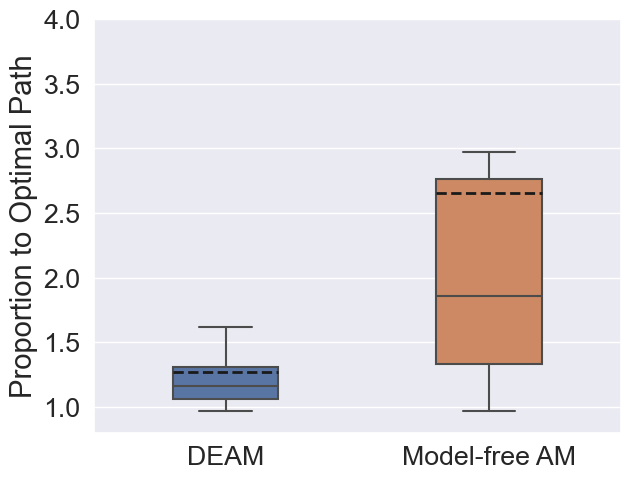}
    \subcaption[]{Path costs (continuous)}
    \label{fig:continuous path_costs}
\end{subfigure}%
\begin{subfigure}[t]{.275\textwidth}
    \centering
    \includegraphics[width=\textwidth]{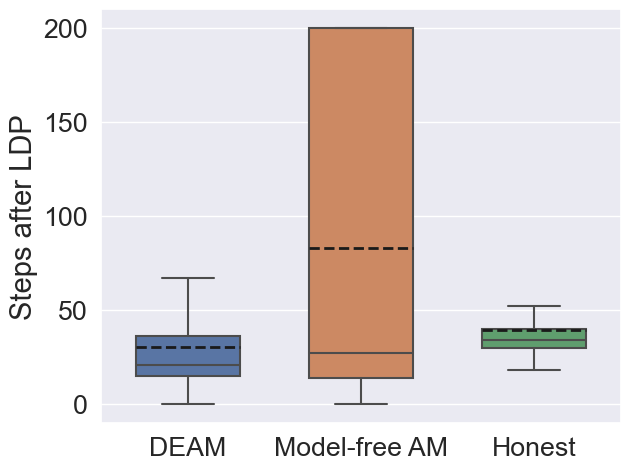}
    \subcaption[]{Steps after LDP (continuous)}
    \label{continuous ldp}
\end{subfigure}%
\caption[]{
Performance: Deceptiveness is measured by the probability of the real goal at different percentages of path completion. Path cost is measured proportional to the optimal, honest path. In the box-plots, the solid lines are medians, dotted lines are means and shaded regions are interquartile ranges.
}
\label{fig:rg probs}
\end{figure*}

\begin{figure}[ht]
\begin{subfigure}[t]{.49\linewidth}
    \centering
    \includegraphics[width=1.\textwidth]{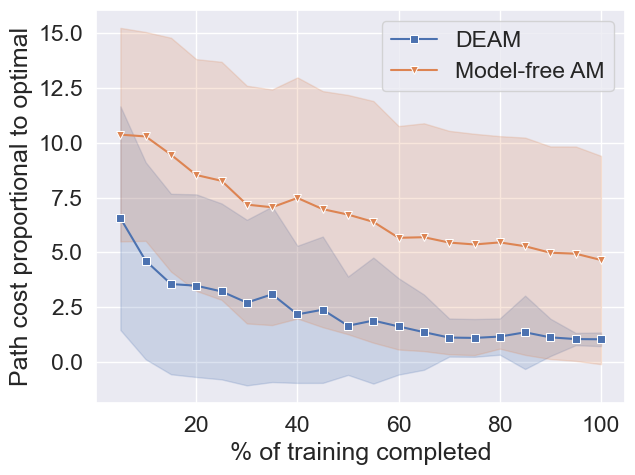}
    \caption{Path cost}
    \label{fig:path-cost-vs-training-time}
\end{subfigure} \hfill%
\begin{subfigure}[t]{.49\linewidth}
    \centering
    \includegraphics[width=1.\linewidth]{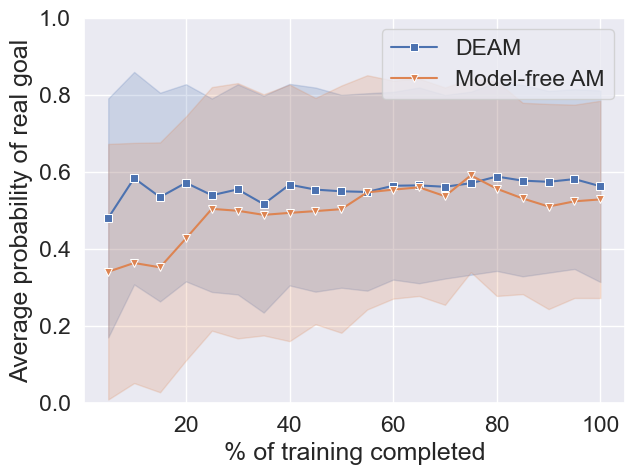}
    \caption{Deceptiveness}
    \label{fig:rg-probs-vs-training-time}
\end{subfigure} \hfill%
\caption[]{Performance vs training interactions. Measures are captured over a trajectory at $20$ intervals during training and averaged over all environment settings. We used $2.4m$ and $3m$ time-steps for $3$ and $5$ goal environments respectively. Shaded regions are a 95\% confidence interval.
}
\label{fig:training-efficiency}
\end{figure}

\textbf{Deceptiveness.}
Figure~\ref{fig:rg probs} shows the trade-off between the deceptiveness and environment rewards.
The real goal probability increases for all agents as they progress along the path, since they move towards and reach the real goal. However, the ambiguity-based agents are consistently more deceptive than the honest agent. We calculated the average real goal probability over each trajectory for all environment settings and compared them across agents using an ANOVA test. We tested assumptions of normality and homoscedasticity using a Shapiro-Wilk and Levene's test respectively. The ANOVA test found a difference in the agent's real goal probability distributions ($p<0.001$). We performed one-sided paired t-tests to test relative deceptiveness. Value iteration AM ($p<0.001$), AM ($p<0.001$) and DEAM ($p<0.001$) all have a lower average probability than the honest agent. None of the ambiguity agents have significantly lower probabilities than each other, indicating that DEAM achieves deception as well as an optimal, model-based AM.

\textbf{Path cost.} The path costs (shown in Figures \ref{fig:discrete path_cost} and \ref{fig:continuous path_costs}) indicate that DEAM is more cost efficient than AM. AM's pre-trained subagents do not learn accurate $Q$-values and policies for states on the deceptive path in the allotted training time, leading to inefficient action selection. In Figure~\ref{fig:AM inefficient action selection}, when AM moves away from the honest paths explored during training, it selects actions that are neither deceptive nor progressive towards the real goal. In other cases, AM moves into unfamiliar states and never reaches the real goal. Since we measure real goal probabilities at different percentages of path completion, it is important to consider it alongside path costs. Due to higher path costs, each path completion percentage corresponds to more steps for AM. Therefore, it travels further with a high real goal probability. This is seen in the LDP results. The average number of steps after the LDP is significantly higher for AM than the other agents. As such, observers have more time between deducing the agent's real goal and the agent arriving there.

\textbf{Continuous environment.} In the continuous environment, AM is the most deceptive 
towards the end of the trajectory. However, this is distorted by scenarios where it fails to reach the real goal, leading to comparatively low probabilities. These scenarios also lead to a high average path cost. Therefore, the apparent deceptiveness is rather a failure to consistently learn to efficiently reach the real goal. 
These results show DEAM's ability to learn more efficiently in complex continuous scenarios due to improved state space exploration. This suggests that DEAM successfully extends dissimulative deception to continuous action space domains.

\textbf{Training efficiency.}
We trained DEAM and AM for $2.4m$ and $3m$ interactions for $3$ and $5$ goal environments respectively. At $20$ intervals, we measured the average real goal probability and path cost in each setting. 
In Figure~\ref{fig:path-cost-vs-training-time} DEAM's path costs stabilise in approximately $70\%$ of the allotted interactions, showing convergence to a deceptive path. AM's path cost is approximately $4 \times$ optimal after training. The constantly high variance indicates a failure converge to a consistent path in the allotted interactions. 
Figure~\ref{fig:rg-probs-vs-training-time} shows the average real goal probability. DEAM stabilises after completing $70\%$ of training as it reaches a consistent path. Although AM appears more deceptive, this is distorted by scenarios where it fails to reach the real goal. DEAM balances the trade-off between deceptiveness and reward accumulation with significantly fewer environment interactions.

\subsection{Active Adversary Experiment Design}
\begin{figure}[ht]
\centering
\begin{subfigure}[t]{0.5\linewidth}
    \centering
    \includegraphics[width=\linewidth]{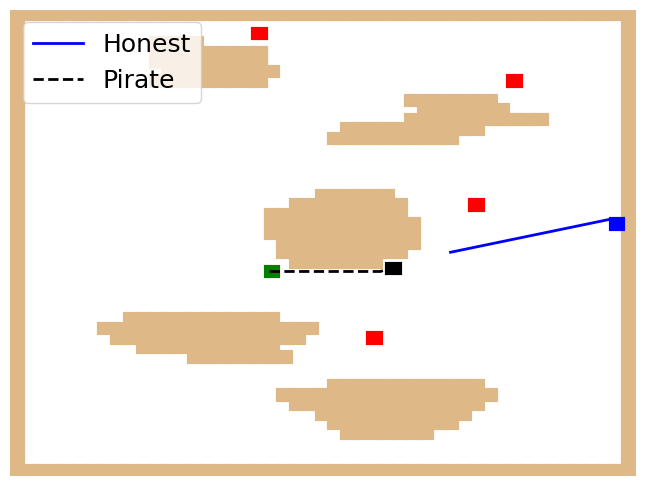}
    \label{fig:pirate examples:honest continuous}
\end{subfigure}%
\begin{subfigure}[t]{0.5\linewidth}
    \centering
    \includegraphics[width=\linewidth]{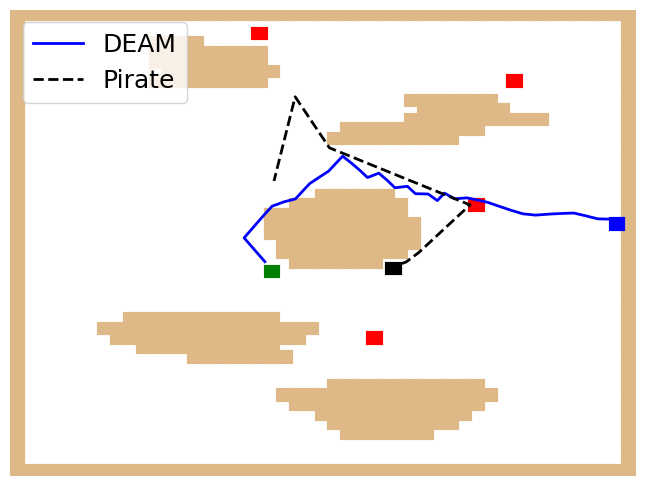}
    \label{fig:pirate examples:ACA continuous}
\end{subfigure}%
\caption{Pirate deception scenario. \textit{Real goal:} Green cell. \textit{Fake goals:} Red cells. \textit{Pirate start:} Black cell. \textit{Agent start:} Blue cell. \textit{Pirate path:} Black line. \textit{Agent path:} Blue line.}
\label{fig:pirate examples}
\end{figure}

We use \citeasnoun{nichols2022adversarial}'s \emph{pirate deception scenario} to assess the agents with an active adversary. The agent must deliver an asset to a goal location as a pirate tries to capture the agent at the goal and steal the asset. This requires deception to lure the pirate away from the real goal, and cost efficiency to prevent being overtaken by the pirate. Figure~\ref{fig:pirate examples} shows an example of the pirate versus an honest agent and DEAM. 
The pirate's path is black and the agent's path is blue. 
The pirate determines the honest agent's real goal early, reducing the problem to a race to the real goal. In contrast, DEAM successfully deceives the pirate towards a fake goal,  allowing DEAM to reach the real goal safely.

The pirate has two components: (1) An intention recognition system to estimate candidate goal probabilities; and (2) A model-based path planner that selects actions towards the most likely goal. The pirate and the agent can only take one action per time-step, with identical action spaces. We use the same environment settings, intention recognition system and agent implementation as Section~\ref{sec:experiment design}. The pirate and agent are initially placed randomly. For each setting, we run $10$ trials with different, random placements, totalling $720$ trials. We use capture rate to evaluate performance.
\subsection{Active Adversary Experiment Results}
\begin{figure}[ht]
\centering
\begin{subfigure}[t]{0.5\linewidth}
    \centering
    \includegraphics[width=\linewidth]{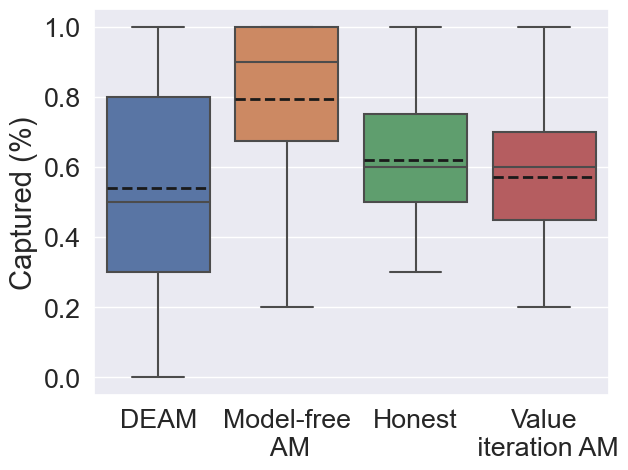}
    \subcaption[]{Discrete}
    \label{fig:pirate performance:discrete}
\end{subfigure}%
\begin{subfigure}[t]{0.5\linewidth}
    \centering
    \includegraphics[width=\linewidth]{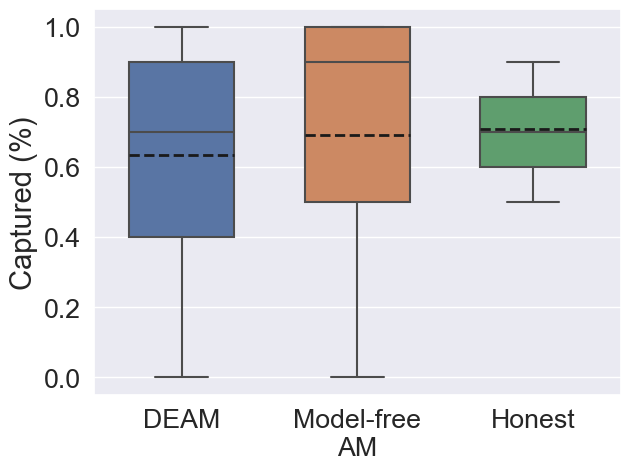}
    \subcaption[]{Continuous}
    \label{fig:pirate performance:continuous}
\end{subfigure}%
\caption{Proportion of captures. For each environment setting, we used 10 random pirate placements, giving a total of 720 runs. Solid lines are medians; dashed lines are means.}
\label{fig:pirate performance}
\end{figure}

Figure~\ref{fig:pirate performance} shows the results. In the discrete domain, DEAM performs best (mean capture rate $= 54.0\%$), followed by value iteration AM (mean capture rate $= 57.1\%$), the honest agent (mean capture rate $= 64.5\%$) and AM (mean capture rate $= 79.5\%$). These results transfer to the continuous domain. AM performs poorly due to learning inefficiencies; as it does not converge effectively, it selects actions that are not deceptive nor progressive towards the real goal. This allows the pirate to overtake it, even when initially deceived.

\begin{figure}[ht]
\centering
\begin{subfigure}[t]{0.5\linewidth}
    \centering
    \includegraphics[width=\linewidth]{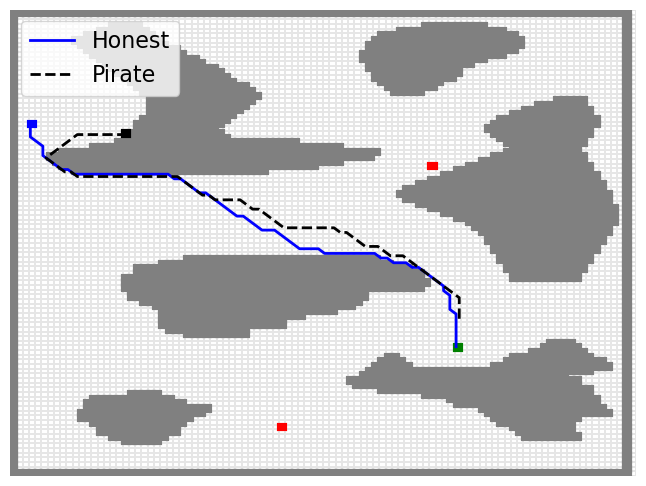}
\end{subfigure}%
\begin{subfigure}[t]{0.5\linewidth}
    \centering
    \includegraphics[width=\linewidth]{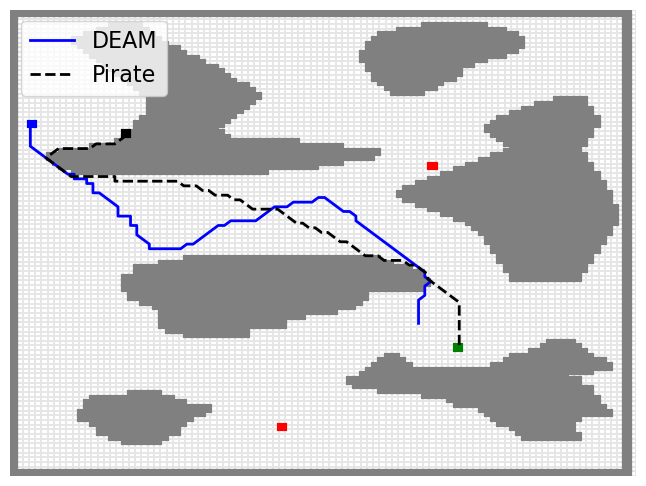}
\end{subfigure}\\%
\begin{subfigure}[t]{0.5\linewidth}
    \centering
    \includegraphics[width=\linewidth]{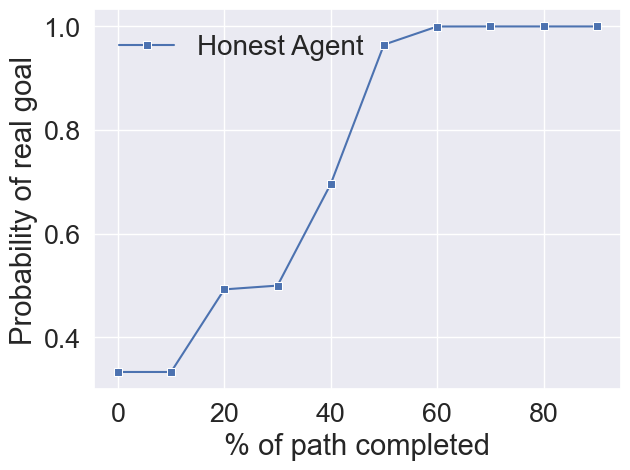}
\end{subfigure}%
\begin{subfigure}[t]{0.5\linewidth}
    \centering
    \includegraphics[width=\linewidth]{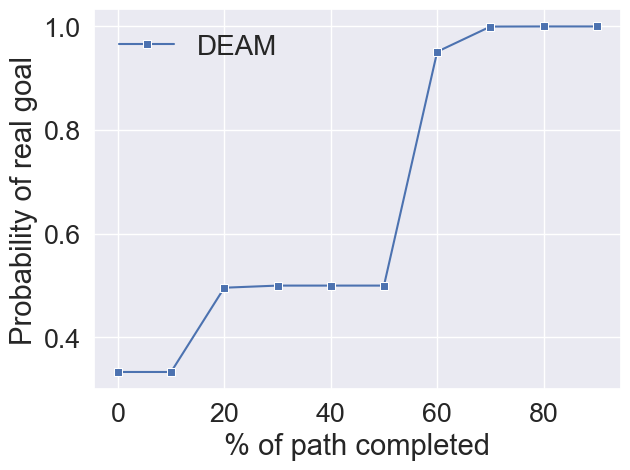}
\end{subfigure}%
\caption{\textit{Counter-productive deception 1:} \textit{Top:} Agent and pirate paths. \textit{Bottom}: Pirate probability estimate. DEAM deceives better than the honest agent, but  DEAM is captured while the honest agent is not. }
\label{fig:counter-productive deception 1}
\end{figure}

\begin{figure}[ht]
\centering
\begin{subfigure}[t]{0.5\linewidth}
    \centering
    \includegraphics[width=\linewidth]{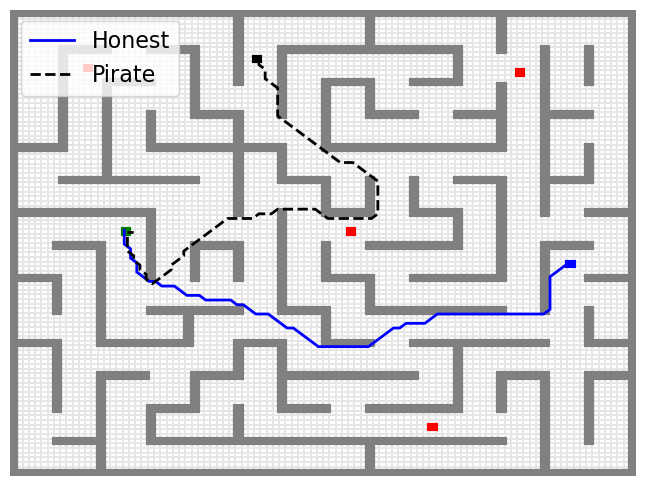}
\end{subfigure}%
\begin{subfigure}[t]{0.5\linewidth}
    \centering
    \includegraphics[width=\linewidth]{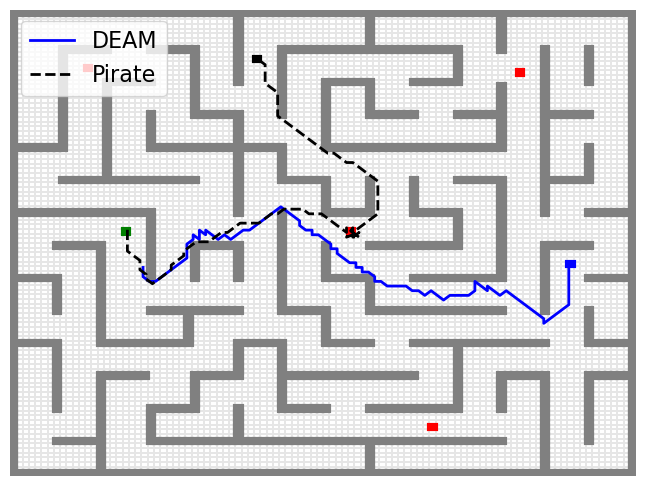}
\end{subfigure}\\%
\begin{subfigure}[t]{0.5\linewidth}
    \centering
    \includegraphics[width=\linewidth]{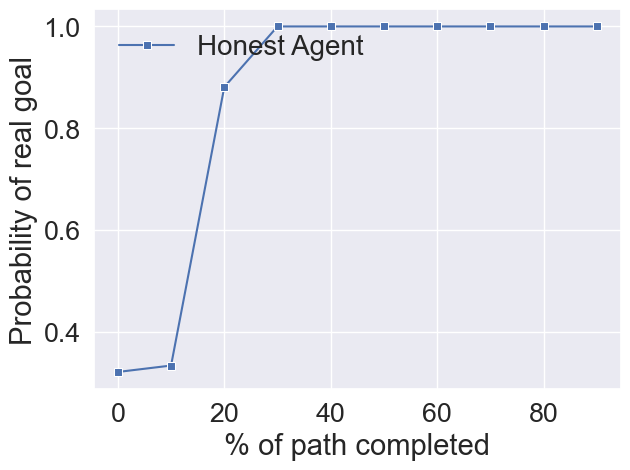}
\end{subfigure}%
\begin{subfigure}[t]{0.5\linewidth}
    \centering
    \includegraphics[width=\linewidth]{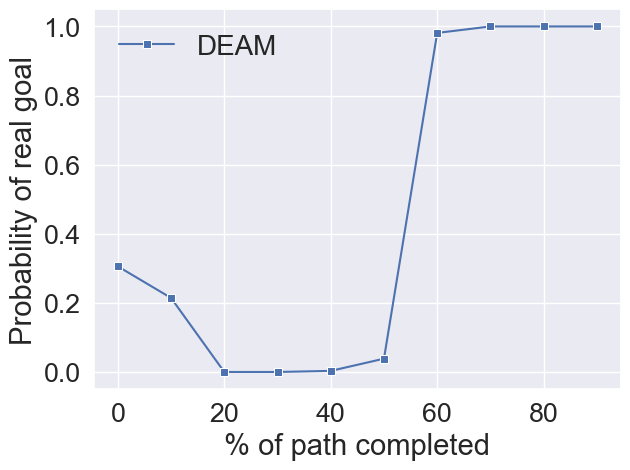}
\end{subfigure}%
\caption{\textit{Counter-productive deception 2:} The deceptive fake goal is along the pirate's path to the real goal. Therefore, DEAM is captured while the honest agent is not.
}
\label{fig:counter-productive deception 2}
\end{figure}

The performance improvement of DEAM and value iteration AM relative to the honest agent is less significant than expected. In some cases, like in Figure~\ref{fig:pirate examples}, DEAM is very effective. However, in other cases, it suffers from counter-productive deceptive actions. Deceptive actions are only effective if they significantly alter the pirate's path away from the real goal. Otherwise, they can harm performance by increasing the path cost. We identify two types of counter-productive actions. First, actions that reduce the pirate's real goal probability, but not enough to change the predicted real goal (see Figure~\ref{fig:counter-productive deception 1}). The pirate's actions matter rather than its real goal probability estimate. Deceptive actions near the end of a trajectory are often in this category as the pirate is already confident in the real goal. Second, actions that alter the pirate's predicted real goal to a fake goal, but the fake goal is on the path to the real goal (see Figure~\ref{fig:counter-productive deception 2}). Despite successfully deceiving the pirate into believing that a fake goal is real, the action is ineffective in sufficiently altering the pirate's path. For deceptive agents to avoid this type of counter-productive action, it needs awareness of the pirate's location, so that it can predict its path and alter it effectively. Neither DEAM nor AM have this awareness by default.

\subsection{Hyperparameter Studies}\label{sec:hyperparameter studies}
DEAM introduces three hyperparameters: a pruning constant $\delta$, a soft-maximum temperature $\tau$ and a soft-maximum temperature decay rate $\lambda_{\tau}$. Here, we investigate the impact of these hyperparameters on performance. We use all $72$ environment settings for each hyperparameter combination. Unless otherwise stated, hyperparameters are set to the values in Table~1 in the Appendix. Figure~\ref{fig:hyperparameter studies} shows the results.

\begin{figure}[htb]
\captionsetup[subfigure]{justification=centering}
\centering
\begin{subfigure}[t]{.45\columnwidth}
    \centering
    \includegraphics[width=1.\textwidth, height=0.91\textwidth]{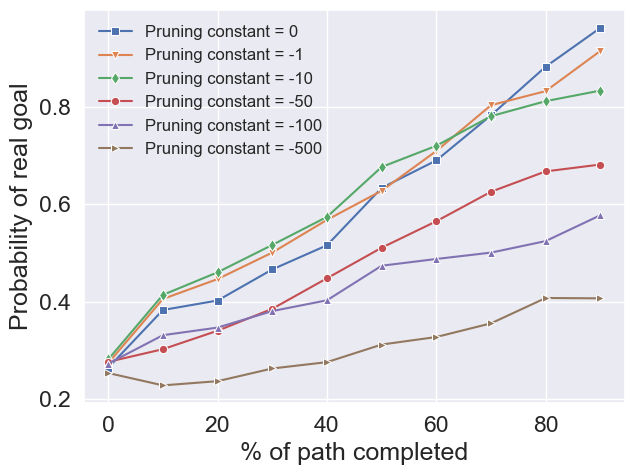}
\end{subfigure}%
\begin{subfigure}[t]{.45\columnwidth}
    \centering
    \includegraphics[width=1.\textwidth, height=0.91\textwidth]{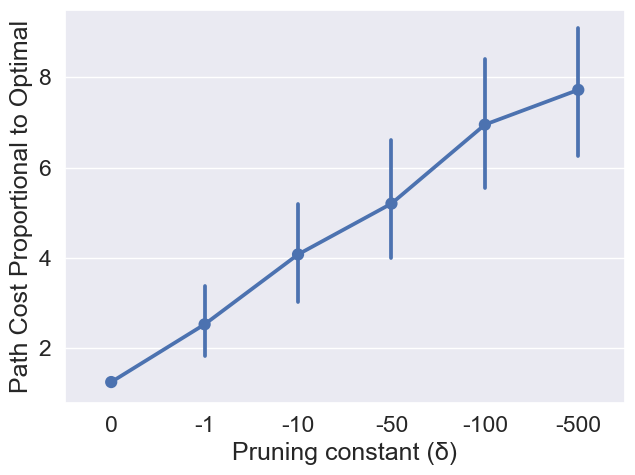}
\end{subfigure}%
\\
\begin{subfigure}[t]{.45\columnwidth}
    \centering
    \includegraphics[width=1.\textwidth, height=0.91\textwidth]{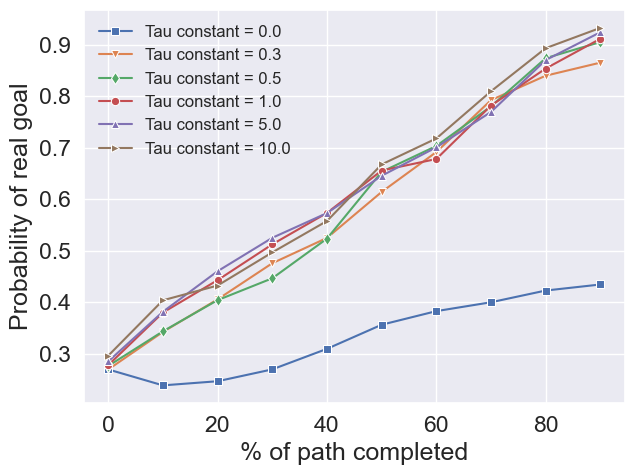}
\end{subfigure}%
\begin{subfigure}[t]{.45\columnwidth}
    \centering
    \includegraphics[width=1.\textwidth, height=0.91\textwidth]{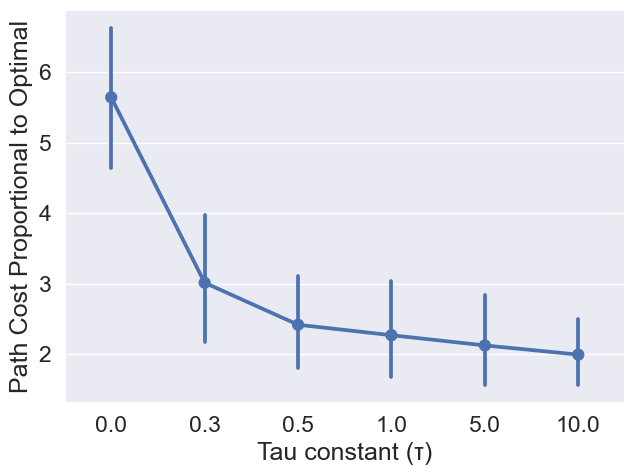}
\end{subfigure}%
\\
\begin{subfigure}[t]{.45\columnwidth}
    \centering
    \includegraphics[width=1.\textwidth, height=0.91\textwidth]{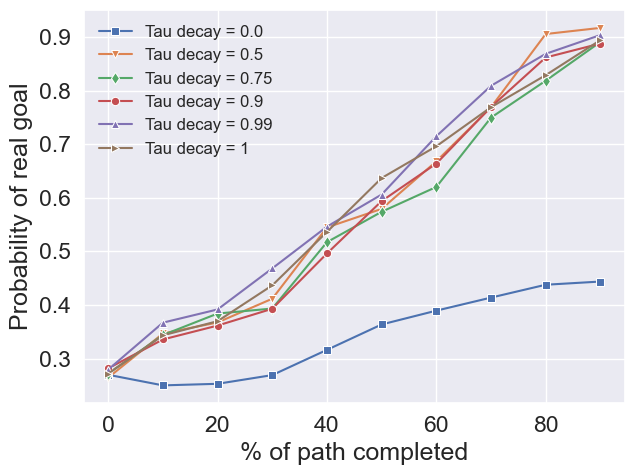}
\end{subfigure}%
\begin{subfigure}[t]{.45\columnwidth}
    \centering
    \includegraphics[width=1.\textwidth, height=0.91\textwidth]{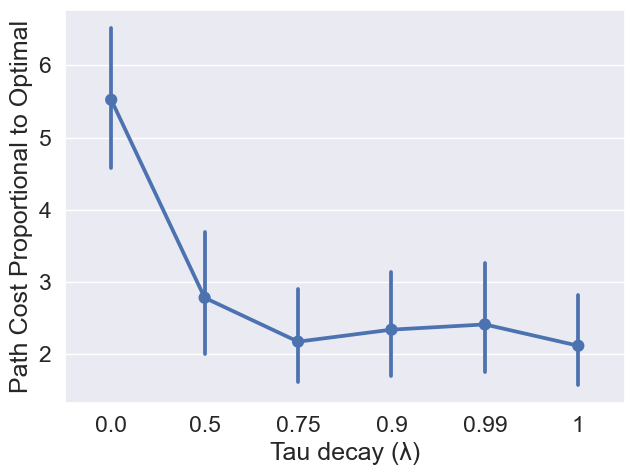}
\end{subfigure}%
\caption[Hyperparameter studies]{DEAM's sensitivity to hyperparameters: pruning constant $\delta$ (top), soft-maximum temperature $\tau$ (middle) and decay rate $\lambda_{\tau}$ (bottom). Left: average real goal probability over the trajectories. Right: average path costs with $95\%$ CI.}
\label{fig:hyperparameter studies}
\end{figure}

\textbf{Pruning constant $\delta$.}
High magnitude $\delta$ values have lower real goal probabilities and higher path costs, suggesting that $\delta$ controls the trade-off between deceptiveness and rewards; when pruning is more aggressive ($\delta \rightarrow 0$) the agent moves more directly to the real goal. However, when $\delta$ is high, DEAM moves ambiguously between candidate goals without reaching the real goal. Therefore, it maintains a low real goal probability throughout the trajectory and has a high path cost. This is undesirable as DEAM only achieves deceptiveness while neglecting rewards. Aggressive pruning is needed to ensure that DEAM reliably reaches the real goal. 

\textbf{Soft-maximum hyperparameters $\tau$ and $\lambda_{\tau}$.} 
$\tau$ controls the stochasticity in the soft-maximum policy; when $\tau \rightarrow 0$, the soft-maximum policy converges to a hard-maximum policy. When $\tau$ starts low or decays quickly, DEAM's behaviour becomes unstable, resulting in high path costs. The added stochasticity in the soft-maximum policy during early training stages enables sufficient exploration of all candidate reward functions, allowing the subagents to learn more accurate $Q$-values before exploiting the deceptive policy as $\tau$ decays. Therefore, a high initial $\tau$ value that decays during training is beneficial for performance. The results also show that decaying $\tau$ slower than $\lambda_{\tau}=0.75$ does not improve performance, indicating that it is not important to maintain high stochasticity in the policy throughout training. This is due to the subagents ability to quickly learn accurate $Q$-values.

\section{Discussion and Broader Impact} \label{sec:discussion and future work}
We introduce DEAM, a dissimulative deceptive RL agent that explores deceptive trajectories during training for application in model-free domains, extends to continuous action spaces and improves training efficiency.
Our results show that DEAM achieves similar performance to an optimal value iteration AM, and extends to a continuous action space environment. Compared to AM, DEAM explores the states that are visited by the deceptive policy, improving deceptiveness, path-costs, and training efficiency.

This study has several limitations. First, DEAM is tied to dissimulative deception, but there are other ways to deceive, such as simulation. Second, we assume that the observer is na\"{i}ve, so is unaware that it is being deceived. Third, we only investigate fully observable environments for the agent and the observer.
Finally, we evaluated only on path planning environments. In future work, we aim to address these limitations by including a more sophisticated observer model and to extend deceptive RL to new types of domains.%


Deceptive AI is a sensitive topic that requires ethical considerations \citep{boden2017principles, danaher2020robot}. \citeasnoun{boden2017principles} argue that robots ``should not be designed in a deceptive way to exploit vulnerable users". A key distinction is the part of the state that is distorted. Deception about the agent's external world is problematic as it distorts the target's perception of reality that exists independent of the deceptive agent \citep{danaher2020robot}. However, deception about the agent's internal state depends on the motivation \citep{danaher2020robot}. DEAM deceives observers about its internal state (reward function) for privacy. This is an important part of responsible AI with positive applications, such as cyber security and enhancing computer-human interaction \citep{adar2013benevolent, dignum2019responsible, sarkadi2019modelling}. Nevertheless, dissimulative deception may be used by malicious agents to deceive vulnerable users. We urge users of our research to consider the ethical implications prior to development.


\bibliography{aaai23}



\newpage
\appendix

\section{Implementation Details}\label{appendex: implementation details}

\subsection{Hyperparameter Choices.}

\begin{table}[htb]
\center
\begin{tabular}{lrr}
\toprule
Parameter                   & Symbol                & Value \\ 
\midrule
Pruning constant            & $\delta$              & 0     \\
soft-max temperature         & $\tau$                & 1     \\
soft-max temperature decay   & $\lambda_{\tau}$      & 0.9 \\
\bottomrule
\end{tabular}
\caption{DEAM hyperparameter choices.}
\label{table:online actor-critic ambiguity hyperparameters}
\end{table}

Table~\ref{table:online actor-critic ambiguity hyperparameters} specifies the hyperparameter choices which are specific to \emph{DEAM}. The hyperparameter studies in Section~\ref{sec:results} discuss these choices. In particular, we found that aggressively pruning leads to the best stability. Although $\delta$ does have control over the trade-off between deceptiveness and reward-maximisation, high magnitude $\delta$ values leads to poor performance as the agent more frequently fails to reach the real goal. Thus, we chose $\delta=0$ since it leads to stable deceptive behaviour.

\subsection{Subagent Implementation.}

\begin{figure}[htb]
\centering
\begin{subfigure}{.49\columnwidth}
    \centering
    \includegraphics[width=1.\textwidth]{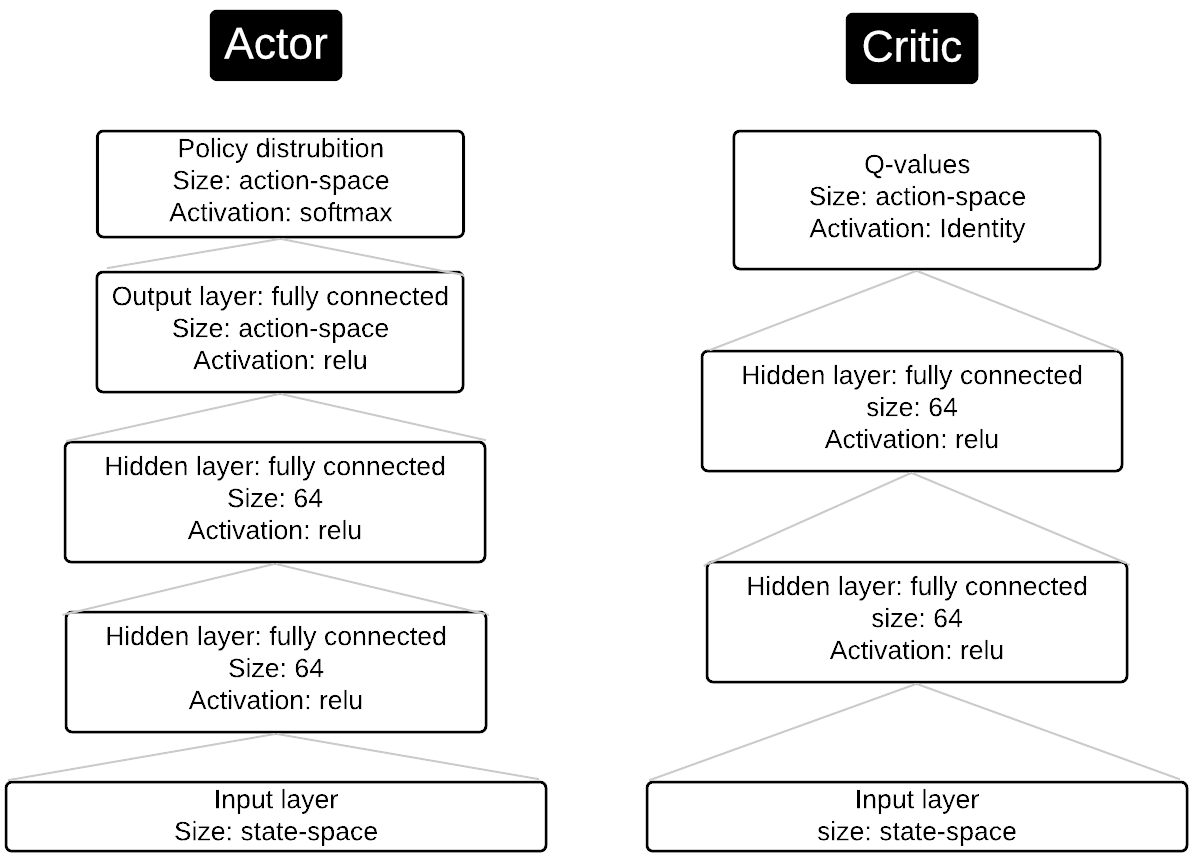}
    \subcaption[]{discrete action space}
    \label{fig:networks:discrete action space} 
\end{subfigure} \hfill%
\begin{subfigure}{.49\columnwidth}
    \centering
    \includegraphics[width=1.\textwidth]{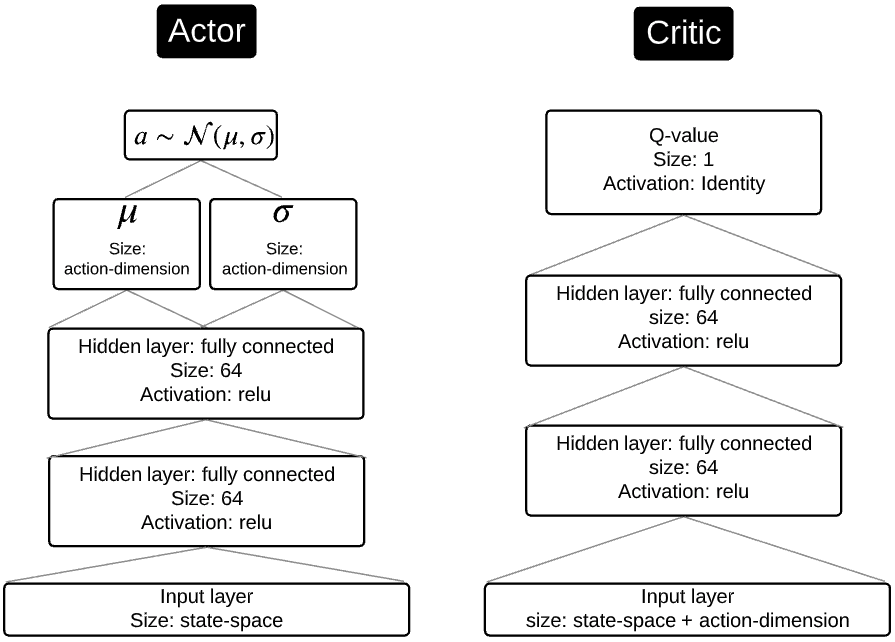}
    \subcaption[]{continuous action space}
    \label{fig:network:continuous action space} 
\end{subfigure}%
\caption[]{The network architectures of the actor and critic MLPs. The same architectures are used for all experiments.}
\label{fig:subagent networks}
\end{figure}

We use SAC as the choice of subagent since it is an \emph{off-policy} AC model. The off-policy characteristic allows subagents to share environment interactions. Table \ref{table:sac hyperparameters} outlines the important hyperparameter settings. Most of the these come directly from the original paper \cite{haarnoja2018soft}, with little tuning for optimal individual performance. We use a fully-connected multi-layer perceptron (MLP) architecture to represent the policies and $Q$-functions. The actor and critic networks do not share parameters. The details of the networks for both continuous and discrete action space environments are shown in Figure~\ref{fig:subagent networks}. For the network architectures and the SAC hyperparameters, we did minimal tuning and kept them consistent across map settings.

\begin{table}[htb]
\center
\begin{tabular}{lrr}
\toprule
Parameter                    & Symbol   & Value       \\ 
\midrule
Horizon                       & $T$      & $4000$      \\
Discount factor               & $\gamma$ & $0.99$      \\
Learning optimisation type    &          & $\textrm{Adam}$        \\
Learning rate                 &          & $3\cdot10^{-4}$   \\
Entropy coefficient (discrete)& $\alpha$ & $0.2$         \\
Entropy coefficient (continuous)& $\alpha$ & $0.01$         \\
Target smoothing coefficient  & $\rho$   & $0.01$     \\
Replay pool size              &          & $10^6$      \\
Minibatch size                &          & $100$       \\ 
\bottomrule
\end{tabular}
\caption{SAC hyperparameter choices used across all environments }
\label{table:sac hyperparameters}
\end{table}

\section{Environment details}

\paragraph{Discrete environment.} 
The discrete environment is an adaptation of MiniGrid \citep{gym_minigrid}. MiniGrid is a lightweight two-dimensional (2D) grid-based environment where an agent needs to navigate from a start location to a goal destination while avoiding any obstacles. We make minor adaptions such that it exactly matches the environment described by \cite{liu2021deceptive}. This allows for comparisons between the two studies. Specifically, we allow diagonal movements and change the reward function to account for distance travelled, rather than a binary reward for reaching the goal. We also adapt the environment to support multiple reward functions/goals in a single environment instance. 
Below we outline and discuss the key components of the environment:
\begin{itemize}
    \item \textbf{State space ($\mathcal{S}$):} At each time-step the agent receives the coordinates of it's current location as the state. This is a tuple of two integers.
    \item \textbf{Action space ($\mathcal{A}$)}: The available actions are:
    \[
    \resizebox{\columnwidth}{!}{%
    \{up, down, left, right, down-left, up-left, up-right, down-right\}
    }
    \]
     This action space is low-dimensional, enabling comparison with AM.
    \item \textbf{Reward functions ($\mathcal{R}$):} The environment gives a vector of rewards at each time-step (one for each candidate reward function $r_i \in \mathcal{R}$) according to the following function:
    \[
    r_i =
    \begin{cases} 
      100 & \textrm{agent reaches the goal} \\
      -1 & \textrm{agent moves laterally}\\
      -\sqrt{2} & \textrm{agent moves diagonally} \\
  \end{cases}
    \]
    The size of the reward vector is therefore the size of the candidate goal set.
\end{itemize}
There are 5 different maps, each with 8 different goal/start-state configurations. We use the same configurations as \citeasnoun{liu2021deceptive}, who use randomly distributed goals and start-states.

\paragraph{Continuous environment.}
The continuous environment is an adaptation of \citeasnoun{henderson2017multitask}'s 2D range-based navigation problem. We change the environment to support multiple reward functions in a single instance. Further, we make a slight change such that the agent cannot pass through walls, rather than penalising them for doing so. We used distance from the goal as the reward at each time-step. This provides a consistent reward signal and thereby helps avoid the cold-start problem. This speeds up learning which allows us to focus on the relative deceptive properties of the agents, rather than the subagents ability to succeed in a sparse reward problem.
\begin{itemize}
    \item \textbf{State space ($\mathcal{S}$):} The agent's current location: $s_t \in \mathbb{R}^2$.
    \item \textbf{Action space ($\mathcal{A}$)}: A continuous direction $\theta \in [0, 2\pi]$ and velocity $v \in [-1, 1]$.
    \item \textbf{Reward functions ($\mathcal{R}$):} For each candidate reward function $r_i \in \mathcal{R}$, the environment gives a reward according to:
    \[
    r_i = - \lVert s_t, g_i \rVert
    \]
    where $g_i$ is the location of the goal corresponding to reward function $r_i$.
\end{itemize}
The action space is continuous. Therefore, value iteration AM cannot learn in this environment.

\section{Additional State Space Exploration Examples}

\begin{figure}[htb]
\centering
\begin{subfigure}[t]{.5\columnwidth }
    \centering
    \includegraphics[width=\textwidth, height=.12\textheight]
    {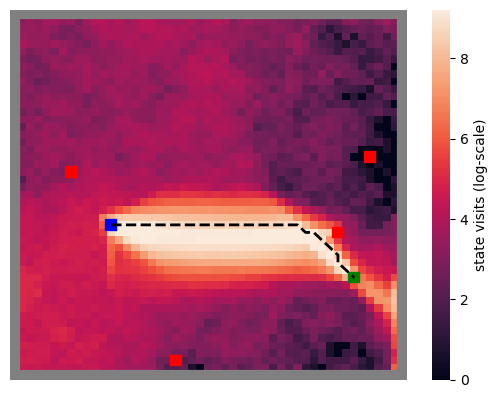}
    \subcaption[]{DEAM: No obstacles}
    \label{DEAM: Map 1}
\end{subfigure}%
\begin{subfigure}[t]{.5\columnwidth }
    \centering
    \includegraphics[width=\textwidth, height=.12\textheight]
    {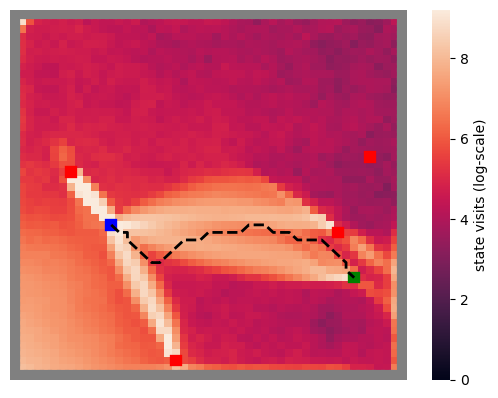}
    \subcaption[]{AM: No obstacles}
    \label{AM: Map 1}
\end{subfigure}%
\\
\begin{subfigure}[t]{.5\columnwidth }
    \centering
    \includegraphics[width=\textwidth, height=.12\textheight]
    {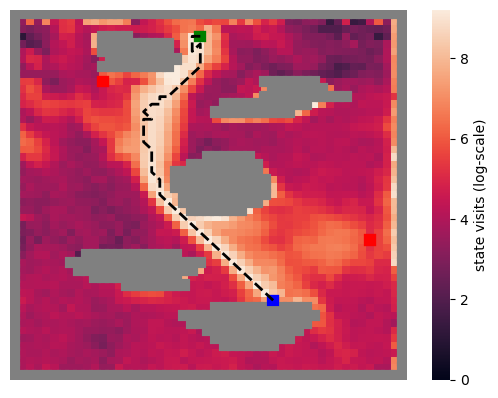}
    \subcaption[]{DEAM: Large obstacles}
    \label{DEAM: Map 2}
\end{subfigure}%
\begin{subfigure}[t]{.5\columnwidth }
    \centering
    \includegraphics[width=\textwidth, height=.12\textheight]
    {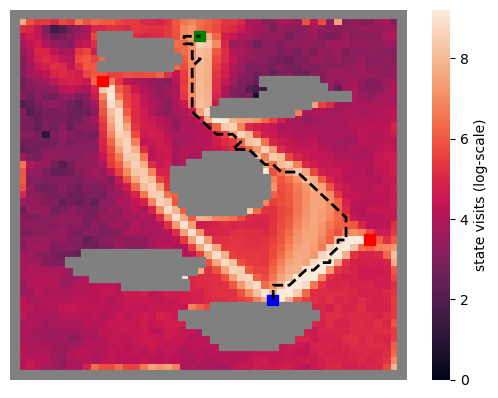}
    \subcaption[]{AM: Large obstacles}
    \label{AM: Map 2}
\end{subfigure}%
\\
\begin{subfigure}[t]{.5\columnwidth }
    \centering
    \includegraphics[width=\textwidth, height=.12\textheight]
    {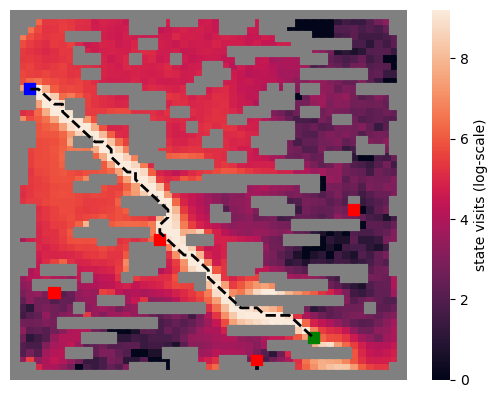}
    \subcaption[]{DEAM: Small obstacles}
    \label{DEAM: Map 3}
\end{subfigure}%
\begin{subfigure}[t]{.5\columnwidth }
    \centering
    \includegraphics[width=\textwidth, height=.12\textheight]
    {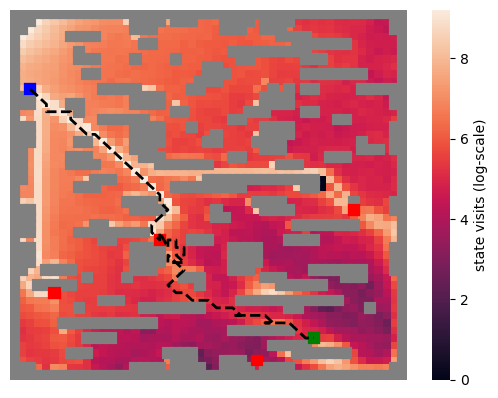}
    \subcaption[]{AM: Small obstacles}
    \label{AM: Map 3}
\end{subfigure}%
\\
\begin{subfigure}[t]{.5\columnwidth }
    \centering
    \includegraphics[width=\textwidth, height=.12\textheight]
    {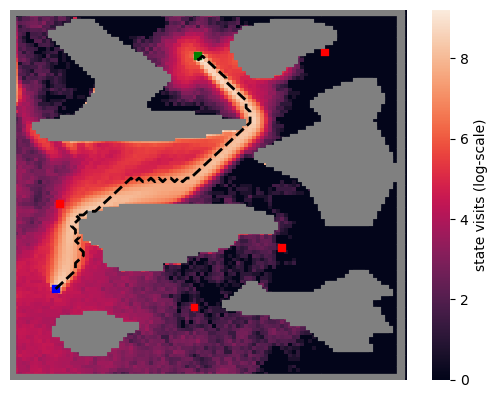}
    \subcaption[]{DEAM: Island}
    \label{DEAM: Map 4}
\end{subfigure}%
\begin{subfigure}[t]{.5\columnwidth }
    \centering
    \includegraphics[width=\textwidth, height=.12\textheight]
    {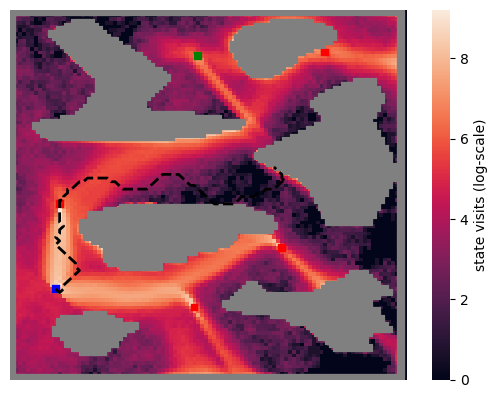}
    \subcaption[]{AM: Island}
    \label{AM: Map 4}
\end{subfigure}%
\\
\begin{subfigure}[t]{.5\columnwidth }
    \centering
    \includegraphics[width=\textwidth, height=.12\textheight]
    {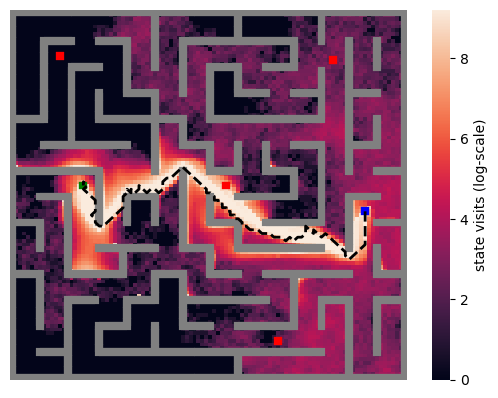}
    \subcaption[]{DEAM: Maze}
    \label{DEAM: Map 5}
\end{subfigure}%
\begin{subfigure}[t]{.5\columnwidth}
    \centering
    \includegraphics[width=\textwidth, height=.12\textheight]
    {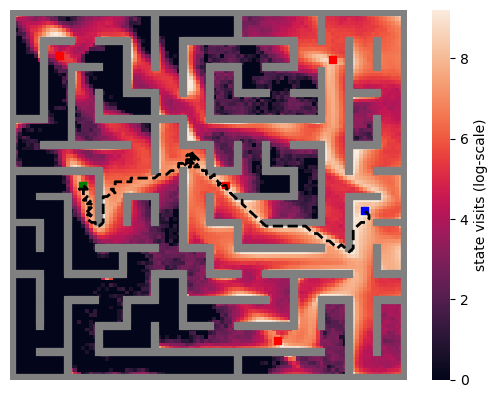}
    \subcaption[]{AM: Maze}
    \label{AM: Map 5}
\end{subfigure}%
\caption[State visitation vs final path (discrete)]{State visitation heatmap vs final deceptive path in each discrete environment map.}
\label{fig:state visitation vs paths (discrete)}
\end{figure}

\begin{figure}[htb]
\centering
\begin{subfigure}[t]{.5\columnwidth }
    \centering
    \includegraphics[width=\textwidth, height=.125\textheight]
    {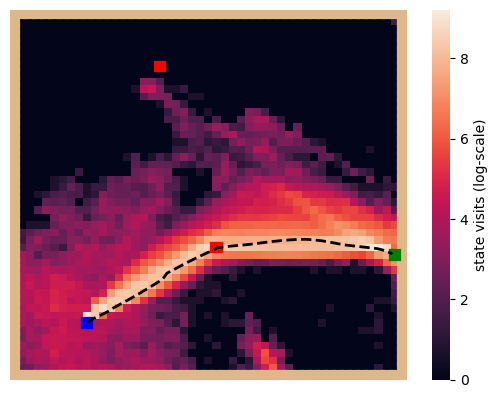}
    \subcaption[]{DEAM: No obstacles}
    \label{Continuous DEAM: Map 1}
\end{subfigure}%
\begin{subfigure}[t]{.5\columnwidth }
    \centering
    \includegraphics[width=\textwidth, height=.125\textheight]
    {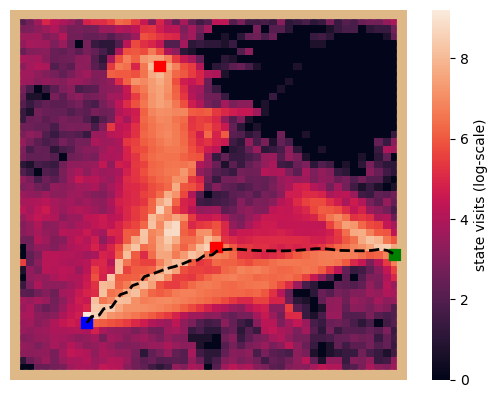}
    \subcaption[]{AM: No obstacles}
    \label{Continuous AM: Map 1}
\end{subfigure}%
\\
\begin{subfigure}[t]{.5\columnwidth }
    \centering
    \includegraphics[width=\textwidth, height=.125\textheight]
    {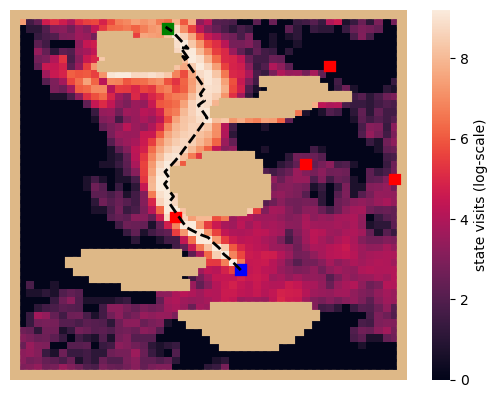}
    \subcaption[]{DEAM: Large obstacles}
    \label{Continuous DEAM: Map 2}
\end{subfigure}%
\begin{subfigure}[t]{.5\columnwidth }
    \centering
    \includegraphics[width=\textwidth, height=.125\textheight]
    {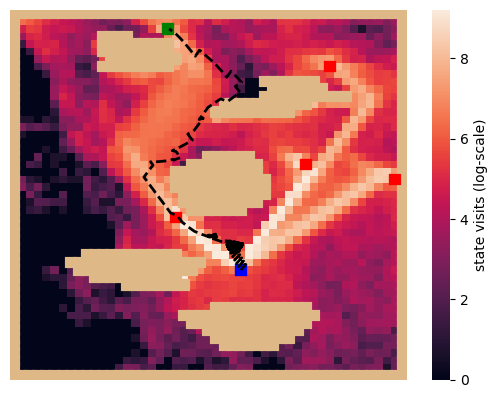}
    \subcaption[]{AM: Large obstacles}
    \label{Continuous AM: Map 2}
\end{subfigure}%
\\
\begin{subfigure}[t]{.5\columnwidth }
    \centering
    \includegraphics[width=\textwidth, height=.125\textheight]
    {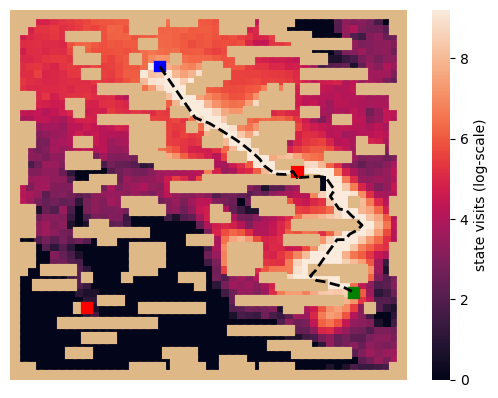}
    \subcaption[]{DEAM: Small obstacles}
    \label{Continuous DEAM: Map 3}
\end{subfigure}%
\begin{subfigure}[t]{.5\columnwidth }
    \centering
    \includegraphics[width=\textwidth, height=.125\textheight]
    {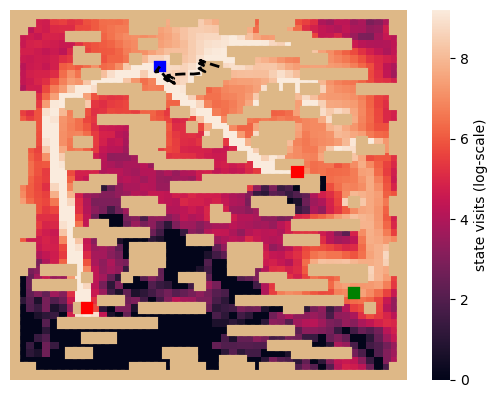}
    \subcaption[]{AM: Small obstacles}
    \label{Continuous AM: Map 3}
\end{subfigure}%
\\
\begin{subfigure}[t]{.5\columnwidth }
    \centering
    \includegraphics[width=\textwidth, height=.125\textheight]
    {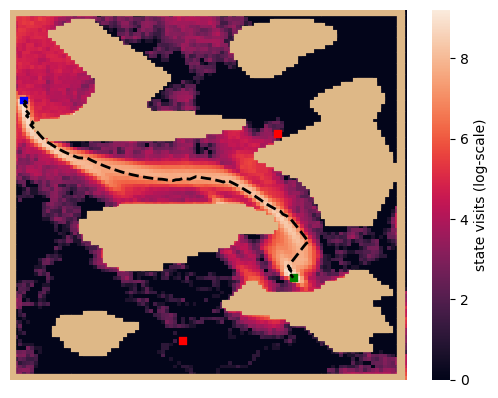}
    \subcaption[]{DEAM: Island}
    \label{Continuous DEAM: Map 4}
\end{subfigure}%
\begin{subfigure}[t]{.5\columnwidth }
    \centering
    \includegraphics[width=\textwidth, height=.125\textheight]
    {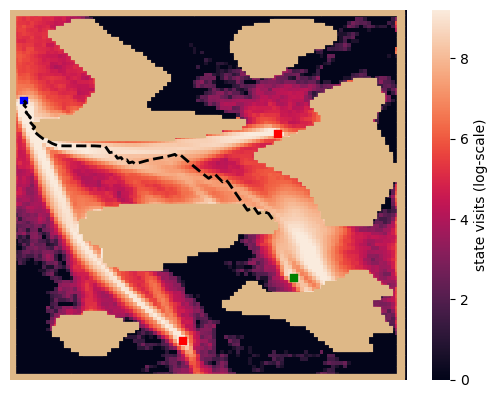}
    \subcaption[]{AM: Island}
    \label{Continuous AM: Map 4}
\end{subfigure}%
\caption[State heatmap vs final path (continuous)]{State visitation heatmap vs final deceptive path in each continuous environment map.}
\label{fig:state visitation vs paths (continuous)}
\end{figure}

Figure~\ref{fig:state visitation vs paths (discrete)} gives further examples of how DEAM (left) and AM (right) explore the state space during training versus their final deceptive path in the discrete environment. Figure~\ref{fig:state visitation vs paths (continuous)} shows the same in the continuous environment. Lighter regions denote more frequently visited states. 
Evidently, DEAM has better overlap between the frequently visited states during training and the final deceptive path, in both the discrete and continuous action space domains. As such, it focuses its training resources on learning a policy and $Q$-values for states that are visited by the deceptive policy, leading to better and more stable performance. This resolves two types of issues that AM has for deceptive RL in model-free domains: (1) Failure to consistently reach the real goal and (2) Inefficient action selection.

\paragraph{Failure to consistently reach the real goal.}
In some cases, AM fails to reach the real goal. This is undesirable since it neglects the reward accumulation component of the dual object. An optimal AM does not have this problem since the pruning procedure ensures that it makes progress to the real goal at each time-step. The issue occurs when AM moves into an unexplored part of the state space. Due to inaccurate $Q$-values for the region, AM is unable to effectively prune actions that move away from the real goal. Figures~\ref{AM: Map 4},~\ref{Continuous AM: Map 3}~and~\ref{Continuous AM: Map 4} show examples of such behaviour. DEAM resolves the issue by exploring deceptive trajectories during training. As such, it learns accurate $Q$-values for the part of that is state space that is visited by the deceptive policy, leading to an effective pruning procedure. Therefore it reliably reaches the real goal in all maps. Figures~\ref{DEAM: Map 4},~\ref{Continuous DEAM: Map 3}~and~\ref{Continuous AM: Map 4} show DEAM's behaviour in the maps where AM fails.

\paragraph{Inefficient action selection.}
Inefficient action selection refers to actions that are neither deceptive nor progressive towards real rewards, and therefore benefit neither component of the dual objective. Figures~\ref{AM: Map 3}~and~\ref{Continuous AM: Map 2} show examples of such behaviour. In Figure~\ref{AM: Map 3}, approximately half way through the trajectory, AM selects actions that move away from the real goal but not in the direction of a fake goal. In Figure~\ref{Continuous AM: Map 2}, at the start of the trajectory, AM wonders in circles until it eventual moves back onto an explored region of the state space. This is slightly hard to see, but is shown by the black smudge slightly above the start state. There are two causes for these inefficient actions: (1) Ineffective pruning like in previous issue and (2) Inaccurate entropy estimates which overstate/understate the deceptiveness of non-deceptive/deceptive actions. This leads to suboptimal action selection. In both cases, the inefficient actions are a result of inaccurate $Q$-values for that part of the state space. Again, DEAM resolves this inefficiency, shown by the behaviour in Figures~\ref{DEAM: Map 3}~and~\ref{Continuous DEAM: Map 2}.

\section{Pruning Constant Decay Results}
\begin{figure}[t]
\centering
\begin{subfigure}[t]{.5\columnwidth}
    \centering
    \includegraphics[width=1.\textwidth, height=0.91\textwidth]{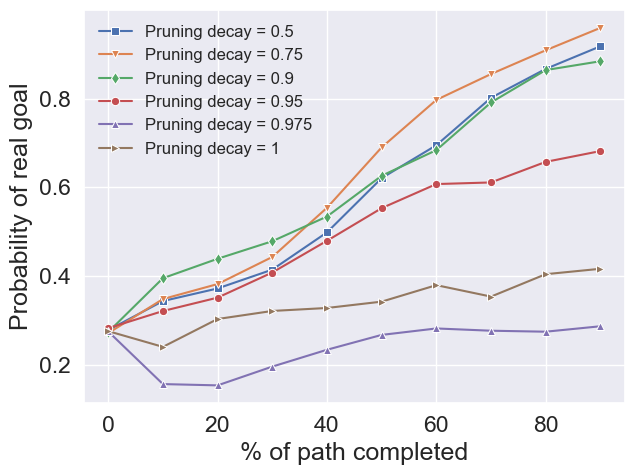}
\end{subfigure}%
\begin{subfigure}[t]{.5\columnwidth}
    \centering
    \includegraphics[width=1.\textwidth, height=0.91\textwidth]{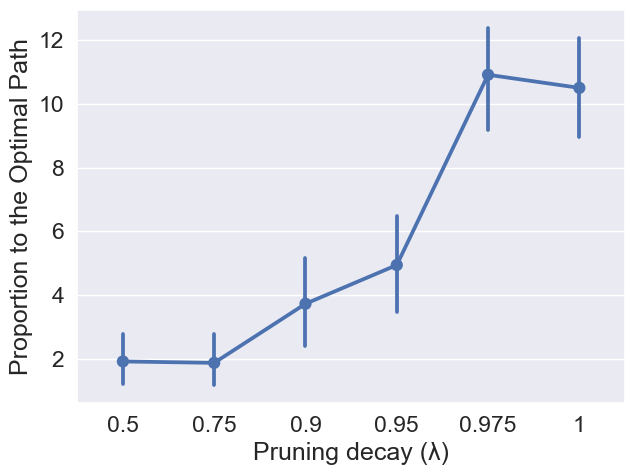}
\end{subfigure} \hfill%
\caption[]{DEAM's sensitivity to a pruning constant decay rate ($\lambda_\tau$) hyperparameter. Left shows the average real goal probability over the trajectories. Right shows the average path costs, surrounded by a $95\%$ confidence interval.}
\label{fig:pruning sensitivity}
\end{figure}

We also experimented with using a decay parameter $\lambda_{\delta}$ to decay $\delta$ during training. This enables high magnitude $\delta$ values, and therefore reduced pruning, during early training stages and low magnitude $\delta$ values, and therefore aggressive pruning, during late training stages. We tried this to encourage state space exploration towards the start of training while moving more directly to the real goal towards the end of training. 
Figure~\ref{fig:pruning sensitivity} shows DEAM's sensitivity to $\lambda_{\delta}$. Left shows the real goal probabilities and right shows the path costs. Like in the pruning constant ($\delta$) hyperparameter results, when $\delta$ is not decayed quickly during training, the agent does not consistently reach the real goal. This behaviour is undesired since it focuses solely on deceptiveness and neglects reward accumulation. Aggressive pruning from the start of training ensures that DEAM reliably reaches the real goal, improving  performance and stability. Therefore, we removed the pruning decay parameter from DEAM.

\section{Statistical Tests and Results}
This section reports the statistical tests and results of agent performance. The mean and standard deviation results for each agent in each domain are reported in Table~\ref{table:mean-std-results}. The best performing agent for each metric is in bold. Here, VI-AM and MF-AM refer to \emph{value iteration AM} and \emph{model-free AM} respectively. Model-free AM has the lowest average real goal probability over a trajectory. However, this is distorted by scenarios where it fails to reach the real goal. This is reflected in model-free AM's path cost, which is significantly higher than the other agents. As expected, the honest agent has the lowest path cost, since it takes an optimal honest path to the real goal. Both value iteration AM and DEAM achieve a good balance between average real goal probabilities and path costs. This is reflected in low number of steps that they take after the LDP. 

\begin{table*}[tb]
\center
\resizebox{\textwidth}{!}{%
\begin{tabular}{lccccccccc}
\toprule
& \multicolumn{3}{c}{\textbf{Average real goal probabilities}} & \multicolumn{3}{c}{\textbf{Path cost}} & \multicolumn{3}{c}{\textbf{Steps after LDP}}\\
\midrule
                        & Discrete               & Continuous            & Both                  & Discrete              & Continuous            &  Both                 & Discrete              & Continuous            &  Both \\
\midrule
\textbf{Honest}          & 0.74 (0.13)           & 0.71 (0.13)           & 0.73 (0.13)           & \textbf{60.6 (31.0)}  & \textbf{50.4 (24.2)}  & \textbf{56.1 (28.6)}  & 29.0 (17.9)           & 39.1 (16.6)           & 34.1 (18.0)\\
\textbf{VI-AM}           & 0.65 (0.13)           & ---                   & ---                   & 69.7 (33.6)           &  ---                  & ---                   & \textbf{21.8 (12.3)}  & ---                   & ---        \\
\textbf{MF-AM}           & \textbf{0.61 (0.23)}  & \textbf{0.57 (0.23)}  & \textbf{0.59 (0.23)}  & 124. (85.2)           & 131. (92.1)           & 127.3 (88.4)          & 80.0 (61.5)           & 83.9 (86.9)           & 82.6 (77.4)\\
\textbf{DEAM}            & 0.62 (0.19)           & 0.58 (0.18)           & 0.60 (0.19)           & 76.1 (44.4)           & 69.5 (54.1)           & 73.2 (49.1)           & 24.9 (20.1)           & \textbf{30.3 (24.7)}  & \textbf{27.6 (22.7)}\\
\bottomrule
\end{tabular}
}
\caption{Mean (standard deviation) of the average real goal probability over a trajectory, path cost and steps after LDP results. VI-AM and MF-AM are value iteration AM and model-free AM respectively. Best performing results are in bold.}
\label{table:mean-std-results}
\end{table*}

To compare the distributions of agent results, we first did a Shapiro-Wilk test (see Table~\ref{table:shapiro-wilk-results}) and Levene's test (see Table~\ref{table:levene's-results}) to verify normality and homoscedasticity respectively. We report path costs proportional to the optimal honest agent as it prevents the data distributions being distorted by the different map sizes. In the case of the path cost and steps after LDP, there is evidence of non-normality ($p<0.05$). Therefore, for these performance measures, we compared distributions using a Kruskal-Wallis H-test. For the average real goal probabilities, compared distributions using an ANOVA test since there is no significant evidence of non-normality or heteroscedasticity. These results are reported in Table~\ref{table:anova-kruskal-results}. For all measures, there is evidence in differences between the agent distributions. As such, we conduct pairwise tests to assess relative performance between agents.

\begin{table*}[tb]
\center
\resizebox{\textwidth}{!}{%
\begin{tabular}{lccccccccc}
\toprule
& \multicolumn{3}{c}{\textbf{Average real goal probabilities}}  & \multicolumn{3}{c}{\textbf{Path cost proportional to optimal}} & \multicolumn{3}{c}{\textbf{Steps after LDP}}\\
\midrule
                & Discrete     & Continuous    & Both         & Discrete     & Continuous    &  Both        & Discrete      & Continuous    &  Both         \\
\midrule
\textbf{Honest} & 0.16 (0.95)  & 0.28 (0.96)   & 0.24 (0.96)  & ---          & ---           & ---           & 0.01 (0.90)  & \textless0.01 (0.84)  & \textless0.01 (0.91)  \\
\textbf{VI-AM}  & 0.28 (0.97)  & ---           & ---          & 0.07 (0.95)  & ---           & ---           & 0.03 (0.92)  & ---           & ---           \\
\textbf{MF-AM}  & 0.11 (0.94)  & 0.10 (0.94)   & 0.05 (0.93)  & \textless0.01 (0.64) & \textless0.01 (0.75)  & \textless0.01 (0.69)  & \textless0.01 (0.40) & \textless0.01 (0.71)  & \textless0.01(0.55)   \\
\textbf{DEAM}   & 0.14 (0.95)  & 0.16 (0.95)   & 0.11 (0.94)  & \textless0.01 (0.67) & \textless0.01 (0.72)  & \textless0.01 (0.70)  & 0.01 (0.90)  & \textless0.01 (0.87)  & \textless0.01 (0.88)  \\
\bottomrule
\end{tabular}
}
\caption{$p$-value ($w$-statistic) from Shapiro-Wilk normality tests.}
\label{table:shapiro-wilk-results}
\end{table*}

\begin{table*}[tb]
\center
\resizebox{\textwidth}{!}{%
\begin{tabular}{ccccccccc}
\toprule
\multicolumn{3}{c}{\textbf{Average real goal probabilities}} & \multicolumn{3}{c}{\textbf{Path cost proportional to optimal}} & \multicolumn{3}{c}{\textbf{Steps after LDP}}\\
\midrule
Discrete    & Continuous    & Both          & Discrete      & Continuous    & Both           & Discrete     & Continuous    &  Both \\
\midrule
0.10 (2.43) & 0.20 (1.70)   & 0.12 (2.48)   & \textless0.01 (10.5)  & \textless0.01 (13.7)  & \textless0.01 (21.33)  & \textless0.01 (8.99) & \textless0.01 (16.64) & \textless0.01 (20.80)\\
\bottomrule
\end{tabular}
}
\caption{$p$-value ($t$-statistic) from Levene's homoscedasticity tests. We include all agents in the discrete test and exclude value iteration AM in the continuous and combined (both) test.}
\label{table:levene's-results}
\end{table*}

\begin{table*}[tb]
\center
\resizebox{\textwidth}{!}{%
\begin{tabular}{ccccccccc}
\toprule
\multicolumn{3}{c}{\textbf{Average real goal probabilities}} & \multicolumn{3}{c}{\textbf{Path Cost}} & \multicolumn{3}{c}{\textbf{Steps after LDP}}\\
\midrule
Discrete    & Continuous    & Both          & Discrete       & Continuous     & Both        & Discrete     & Continuous    &  Both                  \\
\midrule
\textless0.01 (4.55) & \textless0.01 (5.42) & \textless0.01 (11.26) & \textless0.01 (54.07)  & \textless0.01 (15.90)  & 0.03 (4.92) & \textless0.01 (20.97)& \textless0.01 (19.56) & \textless0.01 (38.67)          \\
\bottomrule
\end{tabular}
}
\caption{$p$-value (statistic) from ANOVA and Kruskal-Wallis tests. We used an ANOVA test for the average real goal probability results and a Kruskal-Wallis H-test for the path costs and steps after the LDP results due to evidence of non-normality.}
\label{table:anova-kruskal-results}
\end{table*}

\paragraph{Pairwise tests.} We conducted pairwise tests to compare relative performance between agents, matching samples on the environment settings. For the the real goal probabilities, we did a paired $t$-test. For the path costs and steps after the LDP, we did a Wilcoxon signed rank test since there was evidence of non-normality. Tables~\ref{table:wilcoxin real goal probs},~\ref{table:wilcoxin path costs} and ~\ref{table:wilcoxin LDP} show the results. All the ambiguity agents agents have a lower average real goal probability than the honest agent. None of the ambiguity agents have significantly lower probabilities than the others. These results transfer to the continuous environments for model-free AM and DEAM. The honest agent has a lower path cost than all ambiguity agents. This is expected since it moves directly to the real goal. Neither value iteration AM nor DEAM have lower path costs than each other. However, both have significantly lower costs than model-free AM. This highlights a key contribution of DEAM --- it consistently leads to an efficient deceptive path to the real goal using a model-free learning approach. DEAM takes significantly fewer steps after the LDP than the honest agent and model-free AM. This suggests that the DEAM is more deceptive, since there is a shorter distance between it no longer being deceptive and it reaching the goal. Model-free AM often takes many steps after the LDP since it does not consistently reach the real goal efficiently.

\begin{table*}[t]
\center
\resizebox{\textwidth}{!}{%
\begin{tabular}{lccccccccc}
\toprule
& \multicolumn{3}{c}{\textbf{Honest}} & \multicolumn{3}{c}{\textbf{DEAM}} & \multicolumn{3}{c}{\textbf{MF-AM}}\\
\midrule
                        & Discrete                  & Continuous            & Both                      & Discrete              & Continuous            &  Both                 & Discrete              & Continuous            &  Both                 \\
\midrule
\textbf{VI-AM}           & \textbf{\textless0.01 (6.25)}   & ---                   & ---                       & 0.89 (-1.26)           & ---                   & ---                   & 0.91 (-1.38)           &  ---                  & --- \\
\textbf{MF-AM}           & \textbf{\textless0.01 (3.82)}   & \textbf{\textless0.01 (3.52)}& \textbf{\textless0.01 (8.04)}    & 0.40 (0.25)           & 0.41 (0.25)           & 0.36 (0.36)           & ---                   & ---                   & ---                   \\
\textbf{DEAM}            & \textbf{\textless0.01 (4.45)}   & \textbf{\textless0.01 (4.53)}& \textbf{\textless0.01 (8.35)}    & ---                   & ---                   & ---                   & ---                   & ---                   & ---  \\

\bottomrule
\end{tabular}
}
\caption[Average real goal probability paired $t$-test results]{$p$-value ($t$-statistic) from one-sided paired $t$-tests with the alternative hypothesis that the column agent has a higher average real goal probability over a trajectory than the row agent. For instance, the top left cell tests if the honest agent has a higher average real goal probability than the value iteration AM. Results with $p<0.05$ are in bold.}
\label{table:wilcoxin real goal probs}
\end{table*}

\begin{table*}[t]
\center
\resizebox{\textwidth}{!}{%
\begin{tabular}{lccccccccc}
\toprule
& \multicolumn{3}{c}{\textbf{Honest}} & \multicolumn{3}{c}{\textbf{DEAM}} & \multicolumn{3}{c}{\textbf{MF-AM}}\\
\midrule
                        & Discrete                  & Continuous                & Both                      & Discrete              & Continuous            &  Both                 & Discrete              & Continuous            &  Both                 \\
\midrule
\textbf{VI-AM}           & \textbf{\textless0.01 (0.0)}     & ---                       & ---                       & 0.51 (256.0)          & ---                   & ---                   & \textgreater0.99 (515)          & ---                   & ---                   \\
\textbf{MF-AM}           & \textbf{\textless0.01 (0.0)}     & \textbf{\textless0.01 (0.0)}   & \textbf{\textless0.01 (0.0)}         & \textbf{\textless0.01 (29.0)} & \textbf{\textless0.01 (28)}   & \textbf{\textless0.01 (111.5)}& ----                  & ---                   & ---                   \\
\textbf{DEAM}            & \textbf{\textless0.01 (6.0)}     & \textbf{\textless0.01 (0.0)}   & \textbf{\textless0.01 (0.0)}         & ---                   & ---                   & ----                  & ---                   & ---                   \\

\bottomrule
\end{tabular}
}
\caption[Path cost Wilcoxon signed rank test]{
$p$-value ($w$-statistic) from the Wilcoxon signed rank tests with the alternative hypothesis that the column agent has a lower path cost than the row agent. For instance, the top left cell tests if the honest agent has a lower path cost than the value iteration AM. Results with $p<0.05$ are in bold.}
\label{table:wilcoxin path costs}
\end{table*}

\begin{table*}[t]
\center
\resizebox{\textwidth}{!}{%
\begin{tabular}{lccccccccc}
\toprule
& \multicolumn{3}{c}{\textbf{Honest}} & \multicolumn{3}{c}{\textbf{DEAM}} & \multicolumn{3}{c}{\textbf{MF-AM}}\\
\midrule
                & Discrete               & Continuous            & Both                      & Discrete              & Continuous            &  Both                    & Discrete              & Continuous            &  Both                 \\
\midrule
\textbf{VI-AM}  & \textgreater0.99 (297.5)          & ---                   & ---                       & 0.79 (239.0)          & ---                   & ---                      & \textgreater0.99 (392.0)         & ---                   & ---                   \\
\textbf{MF-AM}  & \textbf{0.03 (132.0)}  & 0.14 (167.0)          & 0.07 (666.5)              & \textbf{\textless0.01 (101.5)}& \textbf{0.01 (88.0)}   & \textbf{\textless0.01 (291.5)}   & ----                  & ---                   & ---                   \\
\textbf{DEAM}   & 0.95 (250.0)           & \textgreater0.99 (267.5)         & \textgreater0.99 (1028.)             & ---                   & ---                   & ----                     & ---                   & ---                   \\

\bottomrule
\end{tabular}
}
\caption[Steps after LDP Wilcoxon signed rank test]{
$p$-value ($w$-statistic) from the Wilcoxon signed rank test with the alternative hypothesis that the column agent takes fewer steps after the LDP than the row agent. For instance, the top left cell tests if the honest agent takes fewer steps after the LDP than value iteration AM. Results with $p<0.05$ are in bold.}
\label{table:wilcoxin LDP}
\end{table*}

\end{document}